\pdfoutput=1

\documentclass[11pt]{article}

\usepackage[final]{acl}

\usepackage{times} 

\usepackage{latexsym}
\usepackage[most]{tcolorbox}
\usepackage{enumitem}
\usepackage{cleveref}
\usepackage{url}
\usepackage{xcolor}
\usepackage{multirow}
\usepackage[subtle]{savetrees}
\usepackage{tocloft}
\usepackage{titletoc}
\usepackage{subcaption}
\usepackage{xspace}
\usepackage[T1]{fontenc}

\usepackage[utf8]{inputenc}

\usepackage{microtype}

\usepackage{inconsolata}

\usepackage{graphicx}
\usepackage{booktabs}
\usepackage{multicol}

\usepackage{booktabs,threeparttable,siunitx,makecell,xcolor}
\newcommand\cavefig{\raisebox{-6pt}{\includegraphics[trim={0em 0em 0em 0em},clip,width=14mm]{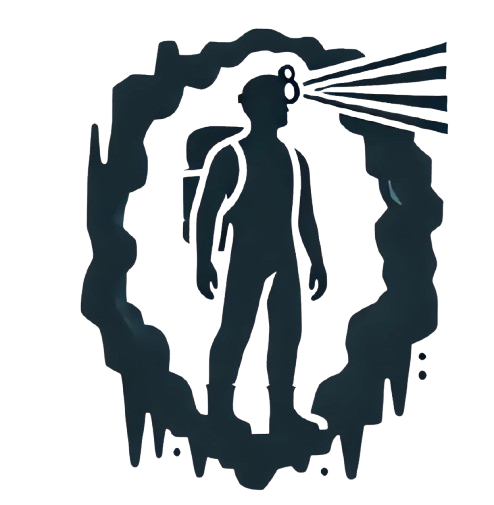}}}

\definecolor{promptbg}{HTML}{F7F7F7} 
\definecolor{promptborder}{HTML}{0077B6} 
\definecolor{prompttitle}{HTML}{023E8A} 
\definecolor{prompttitletext}{HTML}{FFFFFF} 
\usepackage{tocloft}

\newcommand{\ourdata}{CAVE\xspace}

\usepackage{pifont} 
\usepackage{xcolor} 

\newcommand{\cmark}{\textcolor{green}{\ding{51}}} 
\definecolor{darkgreen}{RGB}{0,100,0}
\newcommand{\xmark}{\textcolor{red}{\ding{55}}}

\newcommand{\checkmark}{\textcolor{black}{\ding{51}}} 

\newtcolorbox{promptbox}[1]{
    colback=promptbg, 
    colframe=promptborder, 
    coltitle=prompttitletext, 
    colbacktitle=prompttitle, 
    sharp corners,
    boxrule=1pt, 
    left=10pt, 
    right=10pt, 
    top=6pt, 
    bottom=6pt, 
    fonttitle=\bfseries, 
    title=#1, 
}

\tcbset{
  mysummarybox/.style={
    enhanced jigsaw,   
    breakable,
    colback=gray!10, 
    colframe=black, 
    coltitle=white, 
    boxrule=1pt, 
    left=1mm, 
    right=1mm, 
    top=1mm, 
    bottom=1mm, 
    arc=2mm 
  }
}

\title{\cavefig \ourdata: Detecting and Explaining\\ Commonsense Anomalies in Visual Environments}

\author{
 \textbf{Rishika Bhagwatkar\textsuperscript{1, 2}$^*$},
 \textbf{Syrielle Montariol\textsuperscript{1}$^*$},
 \textbf{Angelika Romanou\textsuperscript{1}},
 \textbf{Beatriz Borges\textsuperscript{1}},
\\
 \textbf{Irina Rish\textsuperscript{2}},
 \textbf{Antoine Bosselut\textsuperscript{1}}
\\
\\
 \textsuperscript{1}EPFL,
 \textsuperscript{2}MILA
\\
 \small{
   \textbf{Correspondence:} \href{mailto:rishika.bhagwatkar@mila.quebec}{rishika.bhagwatkar@mila.quebec}, \href{mailto:syrielle.montariol@epfl.ch}{syrielle.montariol@epfl.ch}
 }
}

\begin{document}
\maketitle
\def\thefootnote{*}\footnotetext{Equal contribution.}\def\thefootnote{\arabic{footnote}}

\begin{abstract}

Humans can naturally identify, reason about, and explain anomalies in their environment. In computer vision, this long-standing challenge remains limited to industrial defects or unrealistic, synthetically generated anomalies, failing to capture the richness and unpredictability of real-world anomalies. In this work, we introduce CAVE, the first benchmark of real-world visual anomalies. 
CAVE supports three open-ended tasks: anomaly description, explanation, and justification; with fine-grained annotations for visual grounding and categorizing anomalies based on their visual manifestations, their complexity, severity, and commonness. These annotations draw inspiration from cognitive science research on how humans identify and resolve anomalies, providing a comprehensive framework for evaluating Vision-Language Models (VLMs) in detecting and understanding anomalies. We show that state-of-the-art VLMs struggle with visual anomaly perception and commonsense reasoning, even with advanced prompting strategies. By offering a realistic and cognitively grounded benchmark, CAVE serves as a valuable resource for advancing research in anomaly detection and commonsense reasoning in VLMs. We release the code and benchmark on our project webpage.\footnote{ \url{https://smontariol.github.io/cave-visual-anomalies/}}

\end{abstract}

\section{Introduction}
\label{sec:intro}

\textit{``If you notice an abnormal situation, please contact an agent.''} Such announcements are commonplace in public spaces worldwide, highlighting a fundamental human trait: the ability to detect anomalies. Identifying uncommon situations, behaviors, and other elements that deviate noticeably from a norm is a natural and expected behavior for humans 
\cite{klein2007data, klein2013seeing}. 

\begin{figure}
    \centering
    \includegraphics[width=\linewidth]{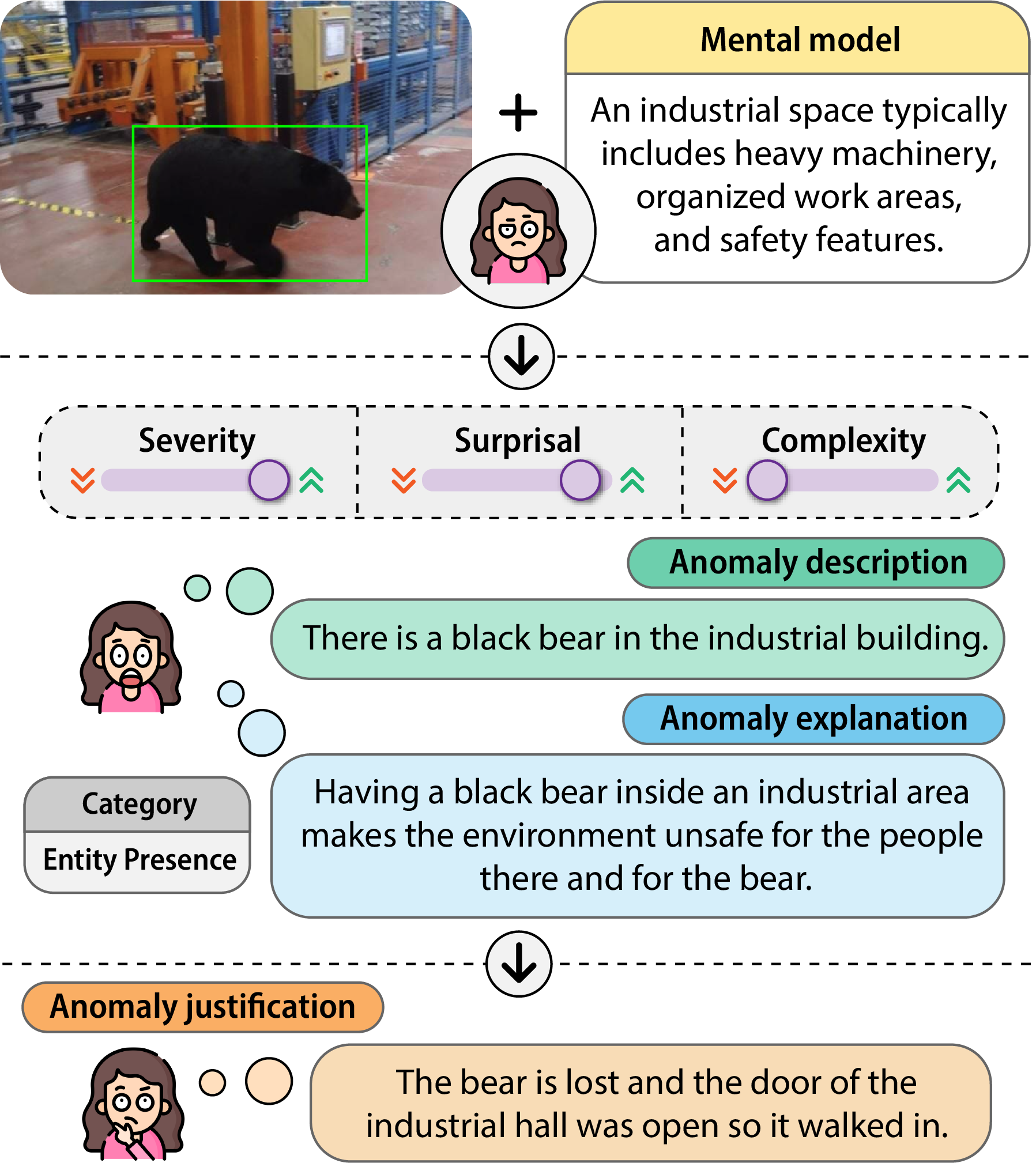}
\caption{\textbf{\ourdata Example.} \ourdata{} contains images captured in real-world scenarios, annotated with commonsense anomalies, along with their textual explanations, justifications and bounding boxes around anomalous elements. It also includes numerical features representing how humans perceive these anomalies.}
\label{fig:example}
\end{figure}

As Vision-Language Models' (VLMs) \cite{li2024llava, openai2024gpt4o, awadalla2023openflamingo, laurenccon2024matters} 
functionality broadens and deployment in real-world scenarios expands~\cite{jin2024llms, xu2024mobility}, so does their exposure to unexpected and novel situations. In this expanding landscape, their ability to differentiate between normal and anomalous situations is crucial for ensuring safe and efficient operation \cite{mullen2024don}.

These models leverage the comprehensive world knowledge and reasoning capabilities of their Large Language Model (LLM) backbones~\citep{mckinzie2024mm1, liu2024visual, karamcheti2024prismatic, laurenccon2024matters}, equipping them to handle a variety of tasks~\cite{liu2024visual, caffagni2024revolution}. However, rare or uncommon situations are inherently underrepresented in training data, making it challenging for models to learn how to recognize them and to react to them. Moreover, VLMs tend to hallucinate image content toward the most probable interpretation, which directly conflicts with their ability to identify unexpected situations \cite{zhou-etal-2023-rome}.

While it is crucial to accurately evaluate VLM's ability to identify and understand anomalous situations, the scope of existing visual anomaly detection benchmarks is limited. In the literature, visual anomaly detection is mainly applied to specific domains such as industrial inspection \cite{chandola2009anomaly, diers2023survey, mvtec, xie2024iad}, medical diagnosis \cite{fernando2021deep, zhang2020covid} or video surveillance \cite{sultani2018real}. 
More recently, commonsense-oriented anomaly detection benchmarks have started to appear. They typically rely on synthetic image generation to create artificial scenarios \cite{visualriddles, li2024nemo, roman-meyer-2024-analysis, whoops, tai2024link}. Non-synthetic approaches rely on domain-specific datasets, such as understanding creative elements in advertisements \cite{malakouti2024benchmarking} or detecting video game glitches \cite{taesiri2024glitchbench}. As a result, existing benchmarks fail to capture the diversity, unpredictability, and realism of real-world anomalies, leaving a critical gap in the evaluation of VLMs' true anomaly detection capabilities.

In this work, we introduce \textbf{C}ommonsense \textbf{A}nomalies in \textbf{V}isual \textbf{E}nvironments (\ourdata), the first visual anomaly benchmark curated from images captured from a human perspective, in real-life settings or as screenshots from smartphones and laptops. Building on top of cognitive science literature, we pair each extracted anomalous image with annotations supporting three open-ended tasks that align with human anomaly detection and sense-making processes: anomaly description, anomaly explanation, and anomaly justification. Additionally, we support an anomaly localization task to evaluate the visual grounding capabilities of the models. We also categorize anomalies based on the type of visual reasoning required to identify them (\textit{e.g.}, spatial or attribute reasoning) and further label them with three numerical features: (a) how severe an anomaly is, (b) how surprising and uncommon it is, and (c) how complex it is to detect (see \Cref{fig:example}). 
\textbf{Our main contributions and findings are as follows}:
\begin{itemize}[topsep=0pt, itemsep=-0.5em, leftmargin=1em]
    \item We propose an anomaly understanding framework that builds upon cognitive science literature regarding the way humans identify and understand anomalies (\Cref{sec:theoretical_framework}). We split the detection process into three sub-tasks formalized as open-ended visual question answering, and include a classification system based on visual manifestations and numerical attributes. This novel framework allows for systematic characterization and annotation of visual commonsense anomalies.
    \item We introduce \ourdata, a benchmark curated from Reddit comprising 361 images designed to evaluate VLMs' ability to detect and understand anomalies (\Cref{sec:dataset}). It captures a wide range of anomalies varying in visual manifestation, commonness, severity, and complexity.
    \item We evaluate 3 proprietary models and 5 open-source state-of-the-art models on \ourdata, experimenting with 5 advanced prompting strategies (\Cref{sec:experiments}). We show that the best model, GPT-4o, only reaches 57\% F1-score on anomaly detection with a multi-step reasoning strategy, highlighting significant room for improvement. 
    \item We analyse VLMs' success and failure modes, finding that they perform better on surprising and severe anomalies but struggle with anomalies involving complex visual perception abilities, especially spatial reasoning and pattern detection.
\end{itemize}


\section{Theoretical Framework}\label{sec:theoretical_framework}

We leverage cognitive science literature to formalize the way humans detect and understand anomalies into a set of tasks. This framework guides our dataset creation process, model assessment, and analysis, allowing us to explore the alignment between human and machine processing of visual anomalies.

\subsection{Perception of the anomaly}

Anomaly detection focuses on identifying deviations from expected patterns \cite{klein2007data, klein2013seeing}. 
In this work, we define an anomaly not simply as a statistical rarity \cite{grubbs1969procedures, chandola2009anomaly, pang2021deep} but as a situation that disrupts an established pattern or expectation. This perspective underscores the key human ability to construct \textit{mental models} of the world and identify deviations from these models~\cite{klein2005problem}. Mental models are cognitive representations of the world, guiding information processing and anticipation of events \cite{borders2024mental}. When an individual encounters something unexpected or surprising that disrupts their established mental model, it can be perceived as an anomaly \cite{klein2007data}; however, humans' accurate perception of situations is limited when doing rapid visual processing~\cite{treisman1982illusory, de1990fixation}. The process of identifying this anomaly depends on three main characteristics.

\noindent \textbf{Anomaly complexity.} Complexity is often operationalized by evaluating perceptual features that influence how quickly and efficiently the brain detects anomalies in visual search tasks \cite{sun2021curious}. Visual anomalies are often salient stimuli that attract attention due to their deviation from expected patterns; more visually complex anomalies require greater cognitive resources for processing \cite{donderi2006visual, guo2023effects}. We leverage this formalization of complexity to assess the difficulty of detecting anomalies in \ourdata.

\noindent \textbf{Anomaly severity.} Anomalies that signal immediate danger or high risk are more likely to be detected. Humans use both cognitive appraisal, \textit{i.e.}, evaluating the potential consequences, and emotional arousal, such as fear and anxiety, to assess the severity of an anomaly \cite{rabeyron2015anomalous}. Hence, we operationalize severity by asking to what extent the anomaly requires immediate action.


\noindent \textbf{Anomaly surprisal.} Surprise-based theories assess severity by how much an event updates prior beliefs (Bayesian Surprise) \cite{itti2009bayesian} or the amount of unexpected information it contains (Information-theoretic Surprise) \cite{baldi2010bits}.
Prediction Error Theory measures surprisal by the magnitude of the discrepancy between expectations and reality, as well as the confidence in the original expectation \cite{friston2005theory}.
Following these concepts, we operationalize surprisal under the question ``How much does the situation deviate from expectations?''.

We use these three formalizations to quantify how humans perceive and detect an anomaly. Similarly to \citet{campbell2024understanding}, we posit that there are commonalities in the way humans and machines process visual information, linking model visual processing limitations with human cognition constraints; and evaluate VLMs' ability to detect anomalies depending on these features.

\begin{figure*}[t!]
    \centering
    \includegraphics[width=\linewidth]{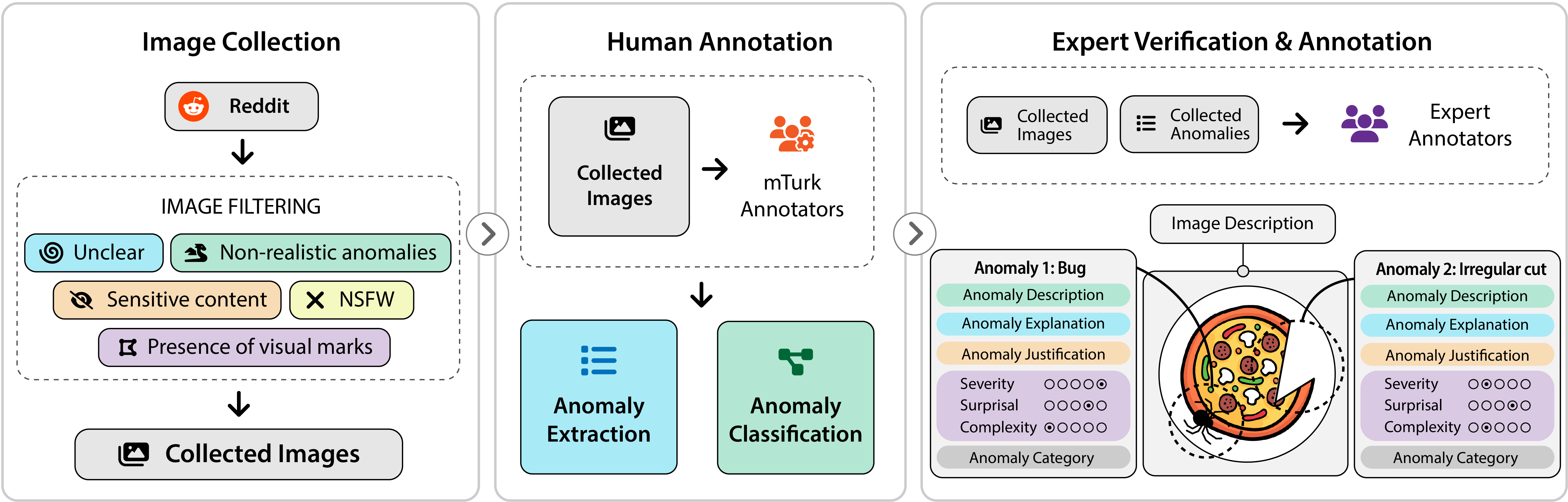}
    \caption{\textbf{An overview of \ourdata{} data collection process.} \textbf{(1) Image Collection}:   Images were sourced from the top 1,000 posts across various subreddits and filtered to ensure high-quality, safe data. \textbf{(2) Human Annotation}: Initial annotations were performed by Mechanical Turk workers, focusing on basic tasks such as anomaly descriptions and anomaly category identification. \textbf{(3)Expert Verification \& Annotation}:  A subsequent round of expert-driven annotation and verification ensured high-quality, consistent annotations across all six tasks, refining and validating the initial labels.}
    \label{fig:cave_process}
\end{figure*}

\subsection{Understanding of the anomaly}

When a human detects an anomaly, the main underlying task is \textbf{anomaly description}~\cite{klein2007data, klein2013seeing}: identifying and articulating what elements in the environment are inconsistent with expectations. 
The next step involves a reassessment of the observer's mental model: why does the situation appear anomalous \cite{klein2023plausibility}? In other words, why did these expectations exist in the first place~\cite{heyes2024rethinking}? We define this intermediate step as \textbf{anomaly explanation}. It aims at assessing the model's understanding of underlying commonsense knowledge on why the situation deviates from the norm. 

Finally, in contrast to typical datasets that often contain artificially generated or staged anomalies, each image in \ourdata represents a real-world scenario captured as an actual photograph or screenshot taken by an individual. These anomalies document real events, prompting the observer to naturally question, ``How did this happen?'' This leads to the final step, termed sense-making~\cite{williams2012explaining, zhang2014towards, klein2023plausibility}, which involves making hypotheses to make sense of the anomaly. To encapsulate this process, we define the \textbf{anomaly justification} task. It involves providing a plausible explanation for the anomaly by describing a sequence of events or circumstances that could have led to the scene.


\subsection{Manifestation of the anomaly}\label{sec:taxonomy}

The manifestation of an anomaly refers to the specific way in which it appears or deviates from the expected pattern in visual data.
This categorization enables targeted benchmarking of VLMs against real-world challenges, ensuring that their anomaly detection capabilities generalize across diverse anomaly manifestations and complementing the cognitive aspect illustrated by the concepts of anomaly complexity, severity and surprisal. 

Inspired by MMBench’s taxonomy of visual reasoning types \cite{liu2025mmbench}, we categorize the ways anomalies manifest in images as follows (see examples for each category in  Appendix \Cref{fig:examples-for-each-category}).
\begin{itemize}[topsep=0pt, itemsep=-0.5em, leftmargin=1em]
    \item \textbf{Entity Presence/Absence}: An object is present in the image when it shouldn’t be, or an expected object is missing.
    \item \textbf{Entity Attribute}: An object exhibits an anomalous attribute, such as an unusual color, shape, label, orientation, or usage.
    \item \textbf{Spatial Relation}: An object is incorrectly positioned or oriented relative to another specific object.
    \item \textbf{Uniformity Breach}: A disruption in an expected pattern, such as an out-of-place element in a uniform or symmetrical arrangement.
    \item \textbf{Textual Anomaly}: Some text in the image conveys an unexpected or contradictory message.
\end{itemize}

\section{Dataset}\label{sec:dataset}

We introduce~\ourdata, a vision-language benchmark which builds on our theoretical framework to evaluate the commonsense anomaly detection and understanding capabilities of VLMs. 

\subsection{Dataset Construction}

\begin{figure*}[ht!]
    \centering
    \includegraphics[width=\linewidth]{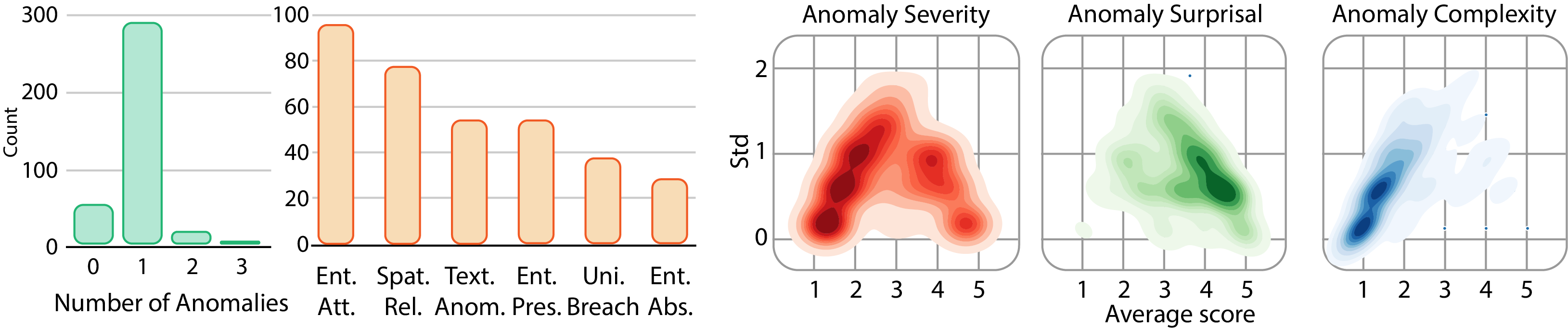}
    \caption{\textbf{\ourdata statistics.} Distribution of the number of anomalies per image (left). Number of images in each anomaly category (middle). Density of severity, surprisal and complexity scores per average score and standard deviation (right).}
    \label{fig:cave_stats_combined}
\end{figure*}

Since our benchmark focuses on real-world, daily-life visual anomalies, our data collection process and annotation strategy are strongly human-centered. 
The dataset creation process is illustrated in \Cref{fig:cave_process}.

\noindent \textbf{Data Collection.} We collect images from four subreddits: \texttt{r/ocdtriggers}, \texttt{r/mildlyconfusing}, \texttt{r/mildlyinfuriating}, and \texttt{r/OSHA}. These subreddits specialize in content featuring unusual or uncommon situations, providing a rich source of real-life anomalies.  

\noindent \textbf{Data Filtering.} We remove images that have unclear content, that contain non-realistic anomalies, and that contain NSFW or sensitive content. We apply automatic and manual filters (see \Cref{supp:data_collection_filtering} for details), and then annotate the remaining images through two annotation rounds. 


\noindent \textbf{Data Annotation.} First, each image was annotated by 5 annotators via Amazon Mechanical Turk. They were asked whether each image was anomalous; If so, they were instructed to (i) describe and explain the anomaly in detail, (ii) describe what they expected instead, and (iii) categorize the anomaly.

Subsequently, expert annotators annotated a single bounding box per anomaly and consolidated the initial textual annotations by validating and formalizing them along the following axes:
\begin{enumerate}[topsep=0pt, itemsep=-0.5em, leftmargin=1em]
    \item \textbf{Anomaly Description (AD):} A textual description of the anomaly in the image.
    \item \textbf{Anomaly Localization (AL):} The coordinates of the bounding box demarcating the anomaly.
    \item \textbf{Anomaly Explanation (AE):} An explanation of why it is anomalous. 
    \item \textbf{Anomaly Justification (AJ):} A realistic and plausible explanation for how the anomaly might have occurred. 
    \item \textbf{Anomaly Category}: Category based on the anomaly manifestation taxonomy outlined in \Cref{sec:taxonomy}. Anomalies about entity attributes, spatial relations, and textual anomalies are the most frequent (\Cref{fig:cave_stats_combined}).
\end{enumerate}

Then, three annotators independently rated each anomaly along the 3 axes:
\begin{itemize}[topsep=0pt, itemsep=-0.5em, leftmargin=1em]
    \item \textbf{Anomaly Severity}: From 1 (does not require action; has no impact on functionality/safety)  to 5 (requires immediate action). 
     \item \textbf{Anomaly Surprisal}: From 1 (common, not very surprising; frequently observed in similar contexts) to 5 (extremely rare).
    \item \textbf{Anomaly Complexity}: From 1 (obvious and easy to notice) to 5 (very hard to detect or requires specific knowledge to identify).
\end{itemize}
\Cref{fig:cave_stats_combined} displays the distribution of these scores. The dataset is skewed toward visually simple anomalies, with severity showing moderate imbalance and surprisal tending toward more unexpected instances, with the latter two having relatively high variance across annotators. A moderate but significant correlation exists between severity and surprisal, with a Spearman correlation of 0.52. This is consistent with the intuition that highly severe anomalies are typically rarer and therefore more surprising. 

We measure the agreement between the 3 annotators (\Cref{tab:numerical_feature_agreement} in \Cref{supp:annotator_agreement}).  
Spearman’s Rank Correlation (0.65) and Krippendorff’s Alpha (0.62) indicate moderate-to-strong agreement among annotators for severity, and weaker for surprisal, which is more subjective. Since complexity and --to a lesser extent--surprisal features have imbalanced distributions, we turn to the more adapted Gwet's AC2 \cite{gwet2008computing}, which shows a much higher agreement for the complexity score (0.76).

In the case of anomaly localization, the bounding box annotations show that most anomalies occupy only a small area of the image, with an average of 24\% (median area is 16\%). 
This illustrates the difficulty of identifying small-sized anomalies in complex real-world scenes. Note that large anomaly areas are mostly associated with textual anomalies, where the entire text region is annotated.

\paragraph{Final dataset.} \ourdata{} consists of \textbf{309 anomalous and 52 normal images} for a total of 361 images. Images have up to 3 anomalies, totaling 334 anomalies, each paired with a unique bounding box. Overall, \ourdata{} exhibits a rich diversity of anomalies (see \Cref{fig:cave_stats_combined} and \Cref{fig:category_per_feature}) across the dimensions of severity, surprisal, complexity and visual manifestation. Moreover, each anomaly is described through our comprehensive multi-task framework, which addresses anomaly detection, explanation, and justification.

\subsection{Evaluation} 

\paragraph{Anomaly Description (AD).}
Since an image can contain multiple anomalies, we evaluate model predictions by performing systematic pairwise comparisons between each ground-truth anomaly description and each model-generated output. To enable scalable evaluation, we employ GPT-4o as a judge to assess whether two descriptions refer to the same anomaly~\cite{liu2024aligning, zheng2023judging, liusie-etal-2024-llm}. This judge achieves 90\% accuracy on 50 manually annotated pairs (GPT-4o vs. human), confirming its reliability (see judge prompt in \Cref{fig:ad_judge_prompt}). Matched pairs are labeled as True Positives (TP), unmatched ground-truth descriptions as False Negatives (FN), and unmatched model outputs as False Positives (FP). We compute the precision, recall, and F1-score using these counts.

\paragraph{Anomaly Localization (AL).}
Given the ground-truth AD for each anomaly, we evaluate a model's ability to predict the correct bounding box by computing the Intersection over Union (IoU) score between the bounding box generated by the model and the ground-truth bounding box. The prompt is given in \Cref{fig:AL_prompt}.

\paragraph{Anomaly Explanation (AE).}
Each AD in \ourdata is paired with a single explanation. We evaluate a model's ability to generate the correct explanation when provided with the ground-truth AD using an LLM judge, comparing model-generated and ground-truth explanations. To validate the judge, we manually label explanation pairs for 50 TPs and FNs from the AD task, achieving 89\% accuracy on the label subset (see judge prompt in \Cref{fig:ae_judge_prompt}).

\paragraph{Anomaly Justification (AJ).}
Since there can be more than one correct anomaly justification, we evaluate a justification quality along three criteria: (1) \textbf{Plausibility}—whether the justification makes sense for the anomaly; (2) \textbf{Relevance}—how well it aligns with the image context; and (3) \textbf{Creativity}—the depth and novelty of the reasoning, beyond generic or trivial explanations. Due to the subjectivity of these criteria, we rely entirely on human evaluation.\footnote{We experimented with LLM-as-a-judge for AJ, but observed low correlation with human assessments, particularly for creativity and plausibility. Hence, we prioritize reliability through human evaluation.} Using the same 50 TPs and FNs as in the AE task, three annotators compare each model-generated justification with the human one and rate it as better, similar or worse. We report the average win rate according to the three annotators.

\begin{table*}[t!]
\centering
\small
\begin{tabular}{lrrrrrr|c}
\toprule
\textbf{Model} & \multicolumn{6}{c|}{\textbf{AD}} & \textbf{AE} \\
\cline{2-7} \cline{8-8}
\addlinespace[0.5em]
 & \textbf{Vanilla} &\textbf{CoT} & \textbf{SoM} & \textbf{CoT + SoM} & \textbf{MS CoT} & \textbf{CoT + consist.} & \textbf{Vanilla} \\

\midrule
Llama3.2 90b        & 24.9    & 36.13 {\scriptsize\textcolor{darkgreen}{(+11.23)}} & 28.00 {\scriptsize\textcolor{darkgreen}{(+3.10)}}  & 29.64 {\scriptsize\textcolor{darkgreen}{(+4.74)}}  & 32.19 {\scriptsize\textcolor{darkgreen}{(+7.29)}}   & 38.56 {\scriptsize\textcolor{darkgreen}{(+13.66)}} & 85.22 \\
LlavaOV 72b         & 27.3    & 27.12 {\scriptsize\textcolor{red}{($-$0.18)}}  & \textbf{43.21} {\scriptsize\textcolor{darkgreen}{(+15.91)}} & 27.11 {\scriptsize\textcolor{red}{($-$0.19)}}  & 29.38 {\scriptsize\textcolor{darkgreen}{(+2.08)}}   & 36.08 {\scriptsize\textcolor{darkgreen}{(+8.78)}} & 85.22 \\
InternVL2.5 38b     & 33.7    & 36.65 {\scriptsize\textcolor{darkgreen}{(+2.95)}}  & 37.79 {\scriptsize\textcolor{darkgreen}{(+4.09)}}  & 33.71 {\scriptsize\textcolor{darkgreen}{(+0.01)}}  & 32.42 {\scriptsize\textcolor{red}{($-$1.28)}}   & 40.00 {\scriptsize\textcolor{darkgreen}{(+6.30)}} & 84.24 \\
QwenVL2.5 72b       & 35.7    & 32.92 {\scriptsize\textcolor{red}{($-$2.78)}}  & 34.33 {\scriptsize\textcolor{red}{($-$1.37)}}  & 29.13 {\scriptsize\textcolor{red}{($-$6.57)}}  & 34.18 {\scriptsize\textcolor{red}{($-$1.52)}}   & 34.32 {\scriptsize\textcolor{red}{($-$1.38)}} & 85.02\\
InternVL2.5 78b     & 36.7    & 39.06 {\scriptsize\textcolor{darkgreen}{(+2.36)}}  & 36.62 {\scriptsize\textcolor{red}{($-$0.08)}}  & 37.55 {\scriptsize\textcolor{darkgreen}{(+0.85)}}  & 35.76 {\scriptsize\textcolor{red}{($-$0.94)}}   & 39.88 {\scriptsize\textcolor{darkgreen}{(+3.18)}} & 83.83 \\
\midrule 
GPT-4o              & \textbf{51.2}    &\textbf{ 54.26} {\scriptsize\textcolor{darkgreen}{(+3.06)}}  & 40.70 {\scriptsize\textcolor{red}{($-$10.50)}} & \textbf{45.05} {\scriptsize\textcolor{red}{($-$6.15)}} & \textbf{56.64} {\scriptsize\textcolor{darkgreen}{(+5.44)}}   & \textbf{53.69} {\scriptsize\textcolor{darkgreen}{(+2.49)}} & 88.04 \\
o1                  & 46.0    & 49.76 {\scriptsize\textcolor{darkgreen}{(+3.76)}}  & 43.54 {\scriptsize\textcolor{red}{($-$2.46)}}  & 41.55 {\scriptsize\textcolor{red}{($-$4.45)}} & 49.50 {\scriptsize\textcolor{darkgreen}{(+3.50)}}   & 52.78 {\scriptsize\textcolor{darkgreen}{(+6.78)}} & \textbf{90.96}\\
Claude & 43.3 & 51.31 {\scriptsize\textcolor{darkgreen}{(+8.00)}} & 34.66 {\scriptsize\textcolor{red}{(-8.65)}} &
43.50 {\scriptsize\textcolor{darkgreen}{(+0.19)}}& 51.31 {\scriptsize\textcolor{darkgreen}{(+8.00)}} & 49.46 {\scriptsize\textcolor{darkgreen}{(+6.15)}} & 80.54\\
\midrule
Average & 37.35 & 40.9 {\scriptsize\textcolor{darkgreen}{(+3.55)}} &  37.35 {\scriptsize\textcolor{darkgreen}{(+0)}} & 35.91 {\scriptsize\textcolor{red}{(-1.44)}} & 40.18 {\scriptsize\textcolor{darkgreen}{(+2.82)}} & 43.10 {\scriptsize\textcolor{darkgreen}{(+5.75)}} & 84.67 \\
\bottomrule
\end{tabular}
\caption{\textbf{AD and AE Results.} F1-scores on the Anomaly Description (AD) task using various prompting strategies (gains over vanilla in parentheses). AE results (last column) are based on the vanilla prompt only.}
\label{tab:combined-results}
\end{table*}

\section{Experiments}\label{sec:experiments}

We evaluate 5 state-of-the-art open-source models and 3 closed-source models (see model details in Appendix \Cref{tab:model_details}) on the AD task (\Cref{sec:ad}), revealing limitations in the form of perception and reasoning errors. To further investigate these shortcomings, we analyze the models’ commonsense reasoning with our two complementary tasks: AE (\Cref{sec:ae}) and AJ (\Cref{sec:aj}). 
Then, we analyze the performance of models on the AD task against numerical features (\Cref{sec:analysis_scores}) and visual manifestation (\Cref{sec:analysis_ac}), identifying the most challenging aspects of anomaly detection.

\subsection{Anomaly Description}\label{sec:ad}
\paragraph{Inference.} We prompt each model to describe the anomalies in the input image and perform evaluation using LLM-as-a-judge. To ensure consistency and reduce evaluation bias, the prompts were carefully aligned with the instructions provided to human annotators (see prompt in \Cref{supp:prompts}). 



\paragraph{Vanilla Prompt Performance.} \Cref{tab:combined-results} shows that using vanilla prompt, the best performance is achieved by GPT-4o with a 51.2\% F1-Score. 

To further understand the limitations of the models, we perform a \textbf{qualitative error analysis}.
We identify two main failure modes with the vanilla prompt. First, \underline{\textit{perception errors}}—hallucinations of missing or non-existent objects, miscounts, or incorrect spatial relations—arise from over-reliance on language priors and weak visual understanding. For instance, in \Cref{fig:ex4}, GPT-4o claims a chair is missing, despite all spots being filled. Second, \underline{\textit{reasoning errors}} occur when models flag contextually normal elements as anomalous due to faulty commonsense reasoning or limited commonsense knowledge. In \Cref{fig:ex1}, QwenVL incorrectly marks a star next to the elevator button “1” as anomalous, overlooking its common use to denote the ground floor. Finally, some cases involve both \underline{\textit{perception and reasoning errors}}. 
Additional examples of model errors can be found in Figures \ref{fig:ex1}-\ref{fig:ex6}.
We perform a manual classification to assign each GPT-4o FP (hallucinated anomaly) into one of these categories in \Cref{tab:count_error_analysis} (first row), finding that around half of them are reasoning mistakes.

\begin{table}[h!]
\centering
\small
\begin{tabular}{lcccc}
\toprule
\textbf{Prompt} & \textbf{Perception} & \textbf{Reasoning} & \textbf{Both} & \textbf{Count}\\
\midrule
Vanilla & 44\% & 49\% & 7\% & 86\\
MS CoT     & 68\% & 32\% & 0 & 95\\
\bottomrule
\end{tabular}
\caption{\textbf{GPT-4o Qualitative Analysis.} Proportion of FP error analysis across prompting strategies as determined by human evaluation.}
\label{tab:count_error_analysis}
\end{table}

\paragraph{Advanced Prompting Strategies.} 
These findings highlight the need for fine-grained visual and contextual reasoning, which general-purpose prompts fail to trigger. As fine-tuning is infeasible due to limited data, we explore five advanced prompting strategies.
\textbf{(1) Chain-of-thought (CoT)}  encourages models to generate explicit reasoning steps before answering \cite{Wei2022ChainOT}; \textbf{(2) Set-of-marks (SoM)} leverages GroundingDINO \cite{Liu2023GroundingDM} to generate object-level visual annotations to enhance visual grounding \cite{yang2023setofmarkpromptingunleashesextraordinary}; \textbf{(3) Combined CoT+SoM} integrates visual grounding annotations to the model's step-by-step reasoning; \textbf{(4) Multi-step CoT} guided models to plan reasoning steps, identify and describe key image elements, and reason step-by-step before answering \cite{xu2025llavacotletvisionlanguage}; \textbf{(5) CoT + Self-consistency} generates three model outputs per image (using temperature=0.5) and applies consensus-based aggregation using the same model, considering only anomalies detected in at least two out of three generations, thereby reducing spurious detections \cite{Wang2022SelfConsistencyIC}.

As shown in \Cref{tab:combined-results}, the five advanced prompting strategies lead to limited improvements over the vanilla baseline across all VLMs (see significance tests in Appendix \Cref{app:AD_results}). CoT + self-consistency demonstrates the strongest overall performance gain (+4.83\%). The highest absolute score remains at 56.64\% with GPT-4o using multi-step reasoning prompting (MS CoT). The performance of many models degrades with the SoM and CoT+SoM prompting strategies; this is due to faulty spatial annotations introducing noise into the anomaly detection task (see example \Cref{fig:som_outputs}). A more fine-grained analysis on TPs and FNs (\Cref{tab:prompt_performance} in appendix) shows that all strategies boost TPs over the vanilla baseline, especially self-consistency and MS CoT, while the number of FNs generally increases with advanced strategies. Notably, CoT, MS CoT, and Self-consistency sharply reduce FP, yielding fewer spurious detections. 

Finally, qualitative analysis (\Cref{tab:count_error_analysis}) shows that for the MS CoT prompting strategy, the false positives predominantly shift toward perception errors (68\% vs. 44\% under vanilla prompting), with fewer reasoning mistakes (32\% vs. 49\%). 
Failure examples are shown in \Cref{supp:failure}. A more detailed analysis of performance across different anomaly categories is provided in \Cref{sec:analysis_ac}. 

\subsection{Anomaly Localization}\label{sec:al}
\paragraph{Inference.} Given the ground truth anomaly description as input, we prompt the best-performing model, GPT-4o, to predict the bounding box coordinates corresponding to the anomaly mentioned in the AD. We then compute the Intersection-over-Union (IoU) score of the predicted bounding box with respect to the ground truth.

\paragraph{Results.} Only 21.7\% of the bounding boxes generated by GPT-4o achieved an IoU $\geq$ 0.10 with the ground truth. The predicted boxes generally under-cover the true anomalies: the mean ratio of predicted-to-ground-truth coverage is 0.69, with 75\% of images exhibiting under-coverage (ratio $<$0.5). Our error analysis shows that localization accuracy drops significantly in cluttered scenes, and that GPT-4o often focuses on smaller subregions rather than the complete anomalous object or context, confirming the quantitative results.

These findings are consistent with prior work~\cite{yang2023dawnlmmspreliminaryexplorations, ramachandran2025doesgpt4ounderstandvision} demonstrating that naive prompting strategies are insufficient for reliable and precise localization with such models. 

\subsection{Anomaly Explanation}\label{sec:ae}
\paragraph{Inference.} 
Building on our findings from the AD task, we examine how well models can explain--rather than identify--visual commonsense anomalies.
For each sample, we prompt the model using the AE prompt in \Cref{supp:prompts}, instructing it to explain why a situation is anomalous, given the ground truth anomaly description.


\paragraph{Results.} 
 All VLMs achieve over 80\% accuracy according to the LLM-as-a-judge evaluation (\Cref{tab:combined-results}). To better understand the link between accurate anomaly description and correct anomaly explanation, we stratify the AE performance results by AD true positive and false negative (see Appendix \Cref{tab:ae-tp-vs-fn-results}).
 The highest-performing models are o1 (93.02\% TP, 88.89\% FN) and GPT-4o (90.86\% TP, 85.22\% FN); on average, there is a 3\% performance gap between TP and FN. Hence, it is slightly easier for models to explain an anomaly that it was able to detect. Overall, this analysis allows us to disambiguate the model's ability to perceive the anomaly in the image from its internal knowledge and understanding of the anomaly. VLMs, despite often possessing the commonsense knowledge required to explain an anomaly, are unable to use that knowledge when performing visual processing. Some of the GPT-4o incorrect AE cases are illustrated in \Cref{supp:failure}, \Cref{fig:ae_error_fig}.

\subsection{Anomaly Justification}\label{sec:aj}
\paragraph{Inference.} We provide the model with both the ground truth anomaly description and explanation alongside the image and ask it to generate a realistic and plausible justification for how the anomaly occurred (see the prompt in \Cref{supp:prompts}).



\begin{figure}[t!]
    \centering
    \includegraphics[width=\linewidth,trim={0 0 0 1cm}, clip]{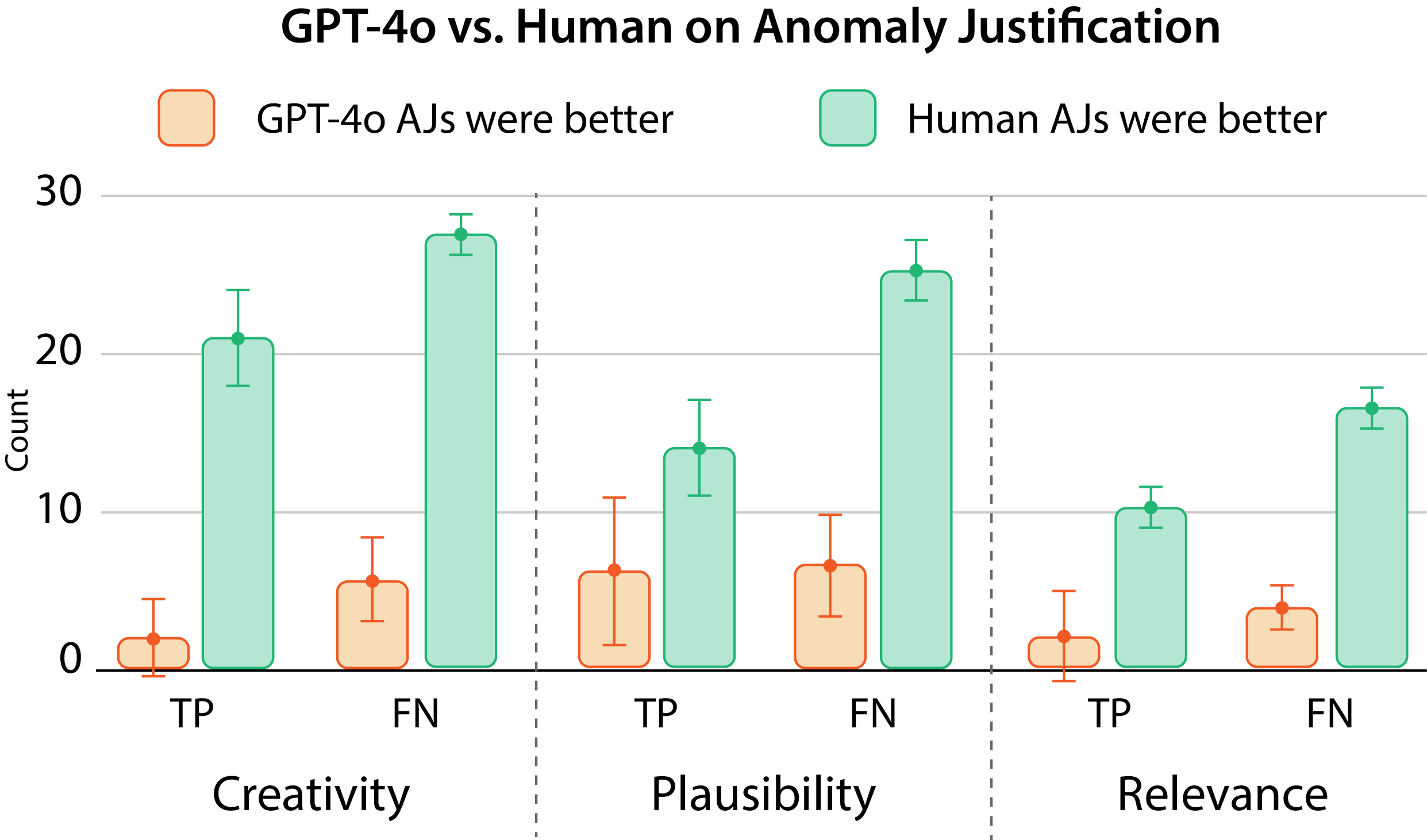}
    \caption{\textbf{AJ Results.} Comparison of GPT-4o vs. Human Anomaly Justifications.} 
    \label{fig:gpt_AJ_eval}
\end{figure}

\paragraph{Results.}
\Cref{fig:gpt_AJ_eval} shows how many GPT-4o-generated justifications are better (orange bars) or worse (green bars) than human justifications for each criterion (plausibility, creativity and relevance), averaged over 3 annotators, along with standard deviation. Among the 50 FN and 50 TP samples, on average, fewer than 7 model-generated justifications outperform humans on any criterion. 
The model’s justifications are less creative, plausible, and contextually relevant when it fails to identify the anomaly (FNs), in line with the AE results. This suggests that these harder cases require both stronger perception and deeper commonsense knowledge to generate plausible explanations. When the model successfully detects the anomaly (TPs), its justifications often resemble human explanations, but lack creativity. GPT-4o tends to favor simpler justifications, attributing anomalies to human forgetfulness, accidental errors, or machine failures, while humans often provide more imaginative explanations, sometimes at the expense of plausibility. This explains the few cases where the model scores higher in plausibility than humans. The model also often generated implausible and irrelevant FNs due to perception and reasoning errors (see examples in \Cref{supp:failure}, \Cref{fig:aj_error_fig}). 
\Cref{fig:internVL_AJ_eval} in \Cref{app:AJ_results} presents results for InternVL, which follows a similar pattern across all features. 

\subsection{Analysis by numerical features}\label{sec:analysis_scores}

\begin{figure}[t]
    \centering
    \includegraphics[width=\linewidth,trim={0 0 0 1cm}, clip]{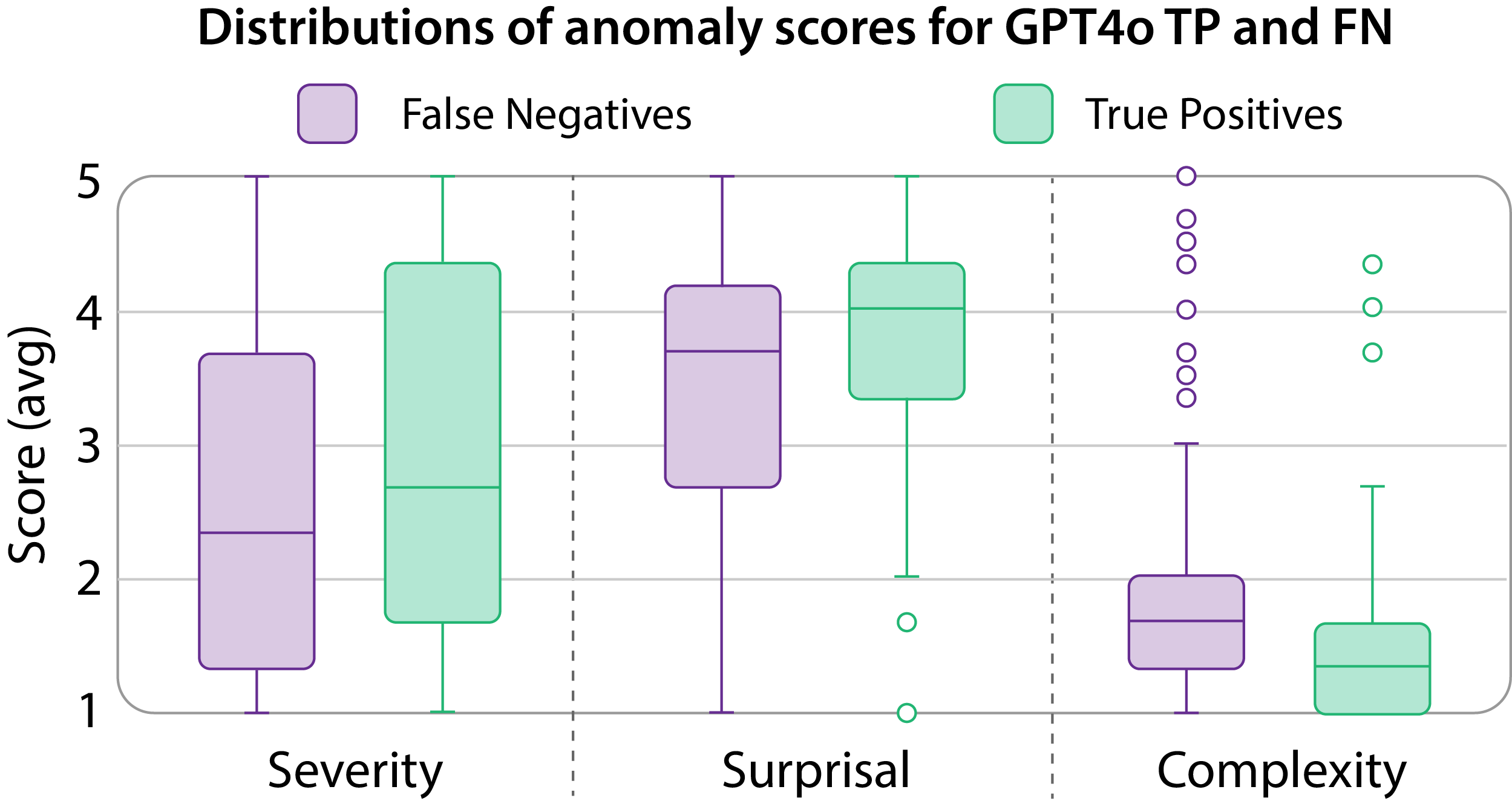}
    \caption{\textbf{Distribution of anomaly descriptions stratified by GPT-4o's TP vs FN} across severity, surprisal, and complexity scores.}
    \label{fig:gpt4o_numerical_features_scores}
\end{figure}

We analyze anomaly detection TPs and FNs across \ourdata's three numerical features: severity, surprisal, and complexity (\Cref{fig:gpt4o_numerical_features_scores}). GPT-4o with vanilla prompt performs best on anomalies that are more surprising and less complex -- \textit{i.e.}, those humans found the most uncommon and easy to spot -- while missed ones are often less severe, less surprising, and more complex. Other models show similar trends (\Cref{app:numerical_features_results}).

\subsection{Analysis by anomaly category}\label{sec:analysis_ac}

Using our anomaly taxonomy (\Cref{sec:taxonomy}), we categorize GPT-4o’s FPs in AD and find it most often hallucinates \textit{attribute, relation}, and \textit{textual anomalies} (\Cref{fig:anomaly-fp}; see classifier details in \Cref{app:AD_results}).
Although \textit{textual anomalies} are among the most frequently hallucinated, they are handled best, with top detection (56.28\%) and strong explanation scores (92.40\%) (See Appendix \Cref{tab:ad_per_category} and \Cref{tab:ae_per_category}). 
In contrast, \textit{uniformity} anomalies, which are rarely hallucinated, are the hardest to detect (28.92\%) and explain (88.28\%).
Interestingly, anomalies on \textit{object absence} show low detection (28.20) but the highest explanation performance (94.52), suggesting models can reason well once the anomaly is identified. Overall, harder-to-detect categories are also harder to explain.

\subsection{Cultural bias assessment}\label{subsec:cultural_bias}

Considering the diversity of cultures and personal experiences, a situation may be perceived as anomalous in one cultural context while appearing entirely normal in another \cite{goto2010cultural, nayak2024benchmarking, ye2023computer}. To investigate this phenomenon, we 
manually annotate which of the anomalies in \ourdata reflect cultural biases. Our analysis shows that while the majority of anomalies are independent from cultural influence, four cases may reflect a Western-centric bias in their annotations. Notably, when GPT-4o is prompted with these images, it consistently identifies them as anomalies, suggesting an implicit alignment with Western cultural norms in the model’s internal knowledge. Further details on the experimental setup and findings are provided in \Cref{supp:cultural_bias}; the four culturally-biased samples are shown in Figure \ref{fig:cultural-biased}.



\section{Related Works}

Historically, image-based anomaly detection has focused on industrial defects such as surface flaws and structural issues \cite{beanTech, mvtec, visA}), tailored for industrial applications and relying on statistical anomaly detection methods. With the rise of VLMs, recent works explore their commonsense reasoning in rare or unusual situations. Many use \textit{synthetic image generation} to create controlled anomalies \cite{visualriddles, wang2024haloquest, tai2024link, li2024nemo, whoops, zhou-etal-2023-rome}, allowing researchers to cover a wide range of scenarios, systematically control the degree and type of abnormality.
These benchmarks typically feature violations of physics or logic, whereas our work uses real-world images -- photographs and screenshots -- with realistic, context-dependent, and varied anomalies. 
The \textit{``in-the-wild''} nature of \ourdata makes anomaly detection significantly more complex than these synthetic datasets, where anomalies are often clearly isolated; real-world anomalies may be subtle, contextually embedded, and require sophisticated perception and reasoning ability to be detected. Benchmarks using real-world unusual situations focus on specific image types, such as creative elements in advertisements \cite{malakouti2024benchmarking} or video game glitches \cite{taesiri2024glitchbench, cao2024physgame}, with limited applicability to commonsense anomalies in real-world images. Additional details about related benchmarks can evaluations can be found in \Cref{app:related_works}, along with a visual comparison of the benchmarks' image types (\Cref{fig:benchmark_comparison}).


In parallel, tailored prompting strategies \cite{xu2025llavacotletvisionlanguage, yang2023setofmarkpromptingunleashesextraordinary} are increasingly designed to tackle vision-language tasks involving complex reasoning, \textit{e.g.}, compositional \cite{thrush2022winoground} or commonsense \cite{zellers2019vcr, bitton2022winogavil}. In this work, we implement several advanced strategies to improve VLMs' perception and reasoning capabilities.

\section{Conclusion and Future Work}

We introduce \ourdata, a multimodal benchmark of 334 visual anomalies in 361 images spanning eight tasks, designed to test VLMs’ real-world anomaly detection and understanding. Leading proprietary and open-source models (>70B parameters) only score $\sim$57 \% F1 on anomaly detection, highlighting significant room for improvement. While they perform better on anomalies seen as highly severe and surprising by humans, they struggle with anomalies that demand complex visual understanding, such as spatial reasoning and detection of pattern violations. Improving anomaly detection requires advances in both visual understanding and commonsense reasoning. Future research could explore fine-grained visual representations for capturing subtle patterns (\textit{e.g.}, uniformity breach) and retrieval-based approaches that leverage large-scale image databases to provide situational commonsense knowledge often missing from existing knowledge sources.


\section{Limitations}

\paragraph{Dataset Size.} Our dataset consists of 361 images and 334 anomalies, which may be considered small compared to large-scale vision-language benchmarks. However, this limitation is counterbalanced by the depth and quality of annotations across seven tasks, including three open-ended tasks (anomaly description, explanation, and justification), one visual-grounding task (anomaly localization) and four classification tasks (anomaly categorization, complexity, severity, and surprisal).  
These factors help mitigate the limited number of examples by offering fine-grained insights into VLMs’ ability to address the tasks. Further, we provide a solid framework for future development and curation.

\paragraph{Dataset Bias.} 
Cultural bias is an inherent challenge in anomaly detection, as what is perceived as anomalous in one culture may be considered normal in another. In our dataset, bias is present due to two main factors. First, image selection bias arises because all images are sourced from Reddit, a platform with a skewed user demographic that does not represent the full diversity of human experiences. Second, anomaly detection and description bias arises from the human annotators, despite the diversity of the backgrounds of the Amazon Mechanical Turk workers and expert annotators. 
Annotation bias occurs both in open-ended tasks (AD, AE and AJ) and numerical ratings—for example, the surprisal score (how uncommon an anomaly is) may vary based on an annotator’s cultural background and personal habits. Moreover, our dataset remains entirely in English, which might further limit its cultural inclusiveness.

To tackle this issue, we implemented several measures.

\noindent\textbf{Diverse annotation team:} We employed a culturally diverse annotation team, with annotators of open-ended tasks coming from 4 different continents. Annotators of numerical features, which are key to represent the subjectivity of an anomaly, come from 3 different countries and cultures. This diversity helps provide a broader perspective on what constitutes an anomaly in different contexts.

\noindent\textbf{Multiple annotations per image:} We had 5 annotators per image during the initial annotation stage, highlight varying perspectives and provide a richer understanding of what different people may consider anomalous; these annotations were later consolidated into the final set of anomalies. Similarly, we had three raters per anomaly for numerical features; these annotations are released as-is, allowing future users of \ourdata to exploit the knowledge stemming from the diversity in numerical scores for each sample.

\noindent\textbf{Inclusive definition of anomaly:} 
We encouraged our annotators to adopt an inclusive definition of anomalies, considering what the majority of people would find anomalous, beyond their own beliefs and expectations. Especially, we encouraged expert annotators to be mindful of potential biases that may influence their perceptions of anomalies, and to consider all original MTurk annotations with an open mind.

\noindent\textbf{Transparency in the collection and annotation process:} 
We provide transparent and comprehensive documentation for the dataset that explains the process followed for collecting and annotating images, allowing future \ourdata users to be aware of the potential bias and coverage limitations of the benchmark.

\paragraph{Dataset Completeness.} 
Our evaluation of anomaly detection relies on precision and recall, and makes the strong assumption that we have exhaustively identified anomalies in each image. This assumption is supported by the extensiveness of our annotation process, with five independent Amazon Mechanical Turk annotations per image and expert validation. Moreover, we excluded ambiguous images where the presence of an anomaly was uncertain or debatable, to minimize borderline cases.

Despite these efforts, it is still possible that some anomalies were overlooked. To account for this, we provide a detailed performance breakdown, reporting the number of false positives, true positives, and false negatives to analyze model behavior in a fine-grained manner.

\paragraph{Dataset Consistency.}
Like most datasets relying on human annotation, ours is subject to errors, subjectivity, and inconsistencies despite extensive efforts in validation and standardization. Differences in individual interpretation could introduce some inconsistencies in open-answer tasks and numerical ratings. 

\paragraph{Model Evaluation.}
We employ LLM-based evaluation as an alternative to costly and time-consuming human assessments. While this enables scalability, it comes with the risk of biases or misjudgments from the LLMs themselves. To address this, we validate LLM-based scores against human annotations and conduct manual evaluations for the two most complex reasoning tasks: anomaly explanation and anomaly justification.

\section{Acknowledgements}

We thank Khurshed P. Fitter, Akshay Kulkarni, Ammar Bhavnagri, Anna Sotnikova, Deniz Bayazit, Silin Gao, Madhur Panwar, Vinay K. Domatoti, Thanmay Jayakumar, Aditya Shirwatkar, and Sepideh Mamooler for the human annotation.

R.B. and I.R. acknowledge support from the Canada CIFAR AI Chair Program and from the Canada Excellence Research Chairs Program.

A.B. also gratefully acknowledges the support of the Swiss National Science Foundation (No. 215390), Innosuisse (PFFS-21-29), the EPFL Center for Imaging, Sony Group Corporation, and a Meta LLM Evaluation Research Grant.

This research was enabled in part by computational resources provided by Mila - Quebec AI Institute.

\bibliography{custom}

\newpage
\onecolumn

\appendix
    
\section*{Appendix Table of Contents}

\addcontentsline{toc}{section}{Appendix}

\setlength{\cftbeforesecskip}{20pt}  
\setlength{\cftbeforesubsecskip}{8pt}  

\setlength{\cftsecindent}{0pt}  
\setlength{\cftsubsecindent}{0pt}  
\setcounter{tocdepth}{2}  
\startcontents[appendix]
\printcontents[appendix]{}{1}{}
\twocolumn

\clearpage
\setcounter{page}{1}

\section{Research Tools}\label{app:compute_details}

\paragraph{Compute details.} 
We evaluated 5 open-source models: InternVL2.5 (38B et 78B parameters) \cite{internvl}, LlavaOneVision (72B) \cite{li2024llava}, QwenVL2.5 (72B) \cite{qwenvl2.5}, and Llama 3.2 (90B) \cite{meta2024llama}.  We use the PyTorch and Hugging Face Transformers implementations for all models examined in this work. Each model is publicly available on the Hugging Face Hub. \Cref{tab:model_details} provides each model's corresponding Hugging Face identifier. All models are run in a zero-shot manner, with a temperature of 0, unless a self-consistency prompting strategy is used. Inference with the large models is done on 4 A100 80B GPUs for up to 3 hours for the full dataset.

\paragraph{Use of AI assistants.} Portions of the code of this paper have been written with the support of a coding assistant (Copilot). All AI-generated codes were thoroughly verified. Portions of the paper were corrected using a writing assistant (Grammarly).


\begin{table*}[h!]
  \centering
  \begin{tabular}{@{} c c @{}}
    \toprule
    Model                 & Identifier                                      \\
    \midrule
    \textit{Open-source Models} \\
    InternVL2.5 38B       & \texttt{OpenGVLab/InternVL2\_5-38B}             \\
    InternVL2.5 78B       & \texttt{OpenGVLab/InternVL2\_5-78B}             \\
    Qwen2.5-VL 72B        & \texttt{Qwen/Qwen2.5-VL-72B-Instruct}           \\
    LlavaOneVision 72B    & \texttt{llava-hf/llava-onevision-qwen2-72b-ov-hf}\\
    Llama3.2 90B Vision    & \texttt{meta-llama/Llama-3.2-90B-Vision}        \\
    \midrule
    \textit{Closed-source Models} \\
    o1                    & \texttt{o1-2024-12-17}                          \\
    GPT-4o                & \texttt{gpt-4o-2024-11-20}                      \\
    Claude                & \texttt{claude-3-5-sonnet-20241022}             \\
    \bottomrule
  \end{tabular}
  \caption{\textbf{Models used.} Overview of the models considered in our study and their corresponding identifiers on the Hugging Face Hub.}
  \label{tab:model_details}
\end{table*}

\section{Related Works}\label{app:related_works}

\subsection{Vision Language Models}

Vision Language Models have made significant progress by integrating powerful vision encoders with LLMs. In most of the models considered in this work (\Cref{tab:model_details}), images are first processed by the vision encoder and then projected into the language model’s embedding space \cite{meta2024llama, qwenvl2.5, internvl}. These visual representations are fused with textual inputs and subsequently passed through the LLM. However, the overall performance of VLMs remains constrained by the capabilities of their vision encoders, particularly in capturing fine-grained visual details or handling out-of-distribution (OOD) images. 

\begin{figure*}
    \centering
    \includegraphics[width=\linewidth]{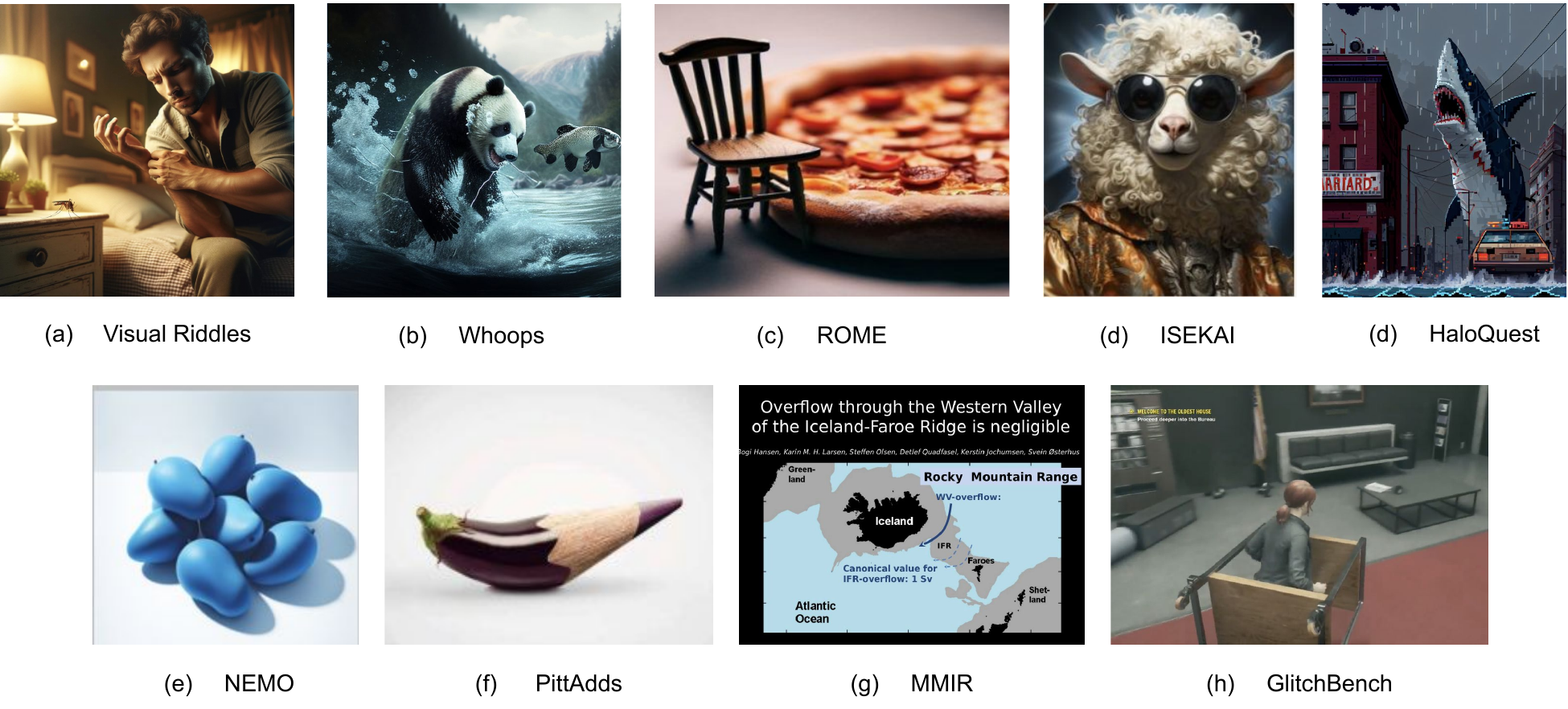}
    \caption{\textbf{Related Benchmark Examples.} Examples of images from related multimodal anomaly-detection benchmarks. More details about each benchmark are given in \Cref{app:related_works} and \Cref{tab:benchmark_comparison}.}
    \label{fig:benchmark_comparison}
\end{figure*}

\subsection{Anomaly detection benchmarks}

Anomaly detection spans various modalities using specialized datasets, from industrial defect identification to autonomous driving \cite{beanTech, mvtec, visA, Bogdoll_2022, survy_visual_ad}. Broadly, anomalies can be classified into \textit{structural} (\textit{e.g.}, physically detectable flaws or distortions in industrial inspections) and \textit{semantic} (deviations at higher hierarchical levels, including the entity, relation, and frame levels) \cite{survy_visual_ad}. In this work, we focus on \textit{semantic anomalies} that necessitate commonsense reasoning for detection and interpretation, hence, we emphasize prior works relevant to this domain.

Several recent multi-modal benchmarks have explored unusual, abstract, or commonsense-defying visual scenarios to evaluate the robustness of VLMs. Visual Riddles \cite{visualriddles} introduces synthetically generated images, each depicting a unique situation and requiring commonsense to answer a question. WHOOPS \cite{whoops} takes a broader approach, generating abnormal images across a wide range of scenarios using three diffusion models. Similar to our work, it extends beyond visual commonsense violations to include anomalies related to social norms, cultural knowledge, and celebrities. The main focus is on explanation generation and image captioning. HaloQuest \cite{wang2024haloquest} attempts to mitigate hallucination by collecting and generating unusual and abstract visual scenes along with VQA designed to trigger hallucinations and use them for VLM fine-tuning. 

Complementing synthetic scenario generation, other benchmarks focus on systematically altering concrete object attributes and relationships to directly probe VLM reasoning. ROME (Reasoning Beyond Commonsense Knowledge) \cite{zhou-etal-2023-rome} explicitly modifies object attributes—such as color, shape, and size—and object relationships using DALL-E 2, creating images that defy commonsense expectations. Similarly, NEMO \cite{li2024nemo} investigates how VLMs recognize objects with uncommon properties, such as a blue mango. ISEKAI \cite{tai2024link} explores a different approach by transferring real-world entities into an alternate world using diffusion models, introducing novel objects and entities and evaluating models on image-pair classification.

A separate line of research focuses on anomalies within structured visual styles, such as advertisements and video games. \citet{malakouti2024benchmarking} leverage the PittAds dataset \cite{hussain2017automatic}, which examines atypical visual elements in advertisements and defines specific tasks like multi-label atypicality classification, atypicality statement retrieval, and atypical object recognition. However, unlike open-ended benchmarks, these tasks constrain atypicality to a specific visual style. Similarly, MMIR \cite{yan2025multimodal} introduces a benchmark to assess VLMs’ ability to detect and reason about semantic mismatches in webpages, presentation slides, and posters—focusing on images where performance is largely driven by OCR capabilities. In contrast, while \ourdata also contains a category for such anomalies, it is limited to a subset of images with less amount of text.

Some recent benchmarks focus on leveraging non-photorealistic yet complex visual environments—such as video games—to evaluate anomaly detection and reasoning. GlitchBench \cite{taesiri2024glitchbench} is a benchmark using unusual and glitched scenes from video games. Similar to ours, one of its strengths is the fact that, since it's not model-generated, there can be many distracting elements in the image, making the detection very challenging. Moreover, it's an open-ended benchmark that is also evaluated using LLMs as a judge. However, all the images are non-realistic and the anomalies defy commonsense. Similarly, PhysGame \cite{cao2024physgame} benchmark models’ ability to identify physical commonsense anomalies in gameplay videos. 

\subsection{Taxonomy-level Comparison}

We also provide a taxonomy-level comparison with synthetic datasets.
Our taxonomy includes six categories (\Cref{sec:taxonomy}): Entity Presence, Entity Absence, Entity Attribute, Spatial Relation, Uniformity Breach and Textual Anomaly. Below we briefly contrast these with existing synthetic benchmarks.

\textbf{Novel Anomaly Manifestations:} Real-world images naturally capture a broader and more open-ended set of anomalies. Entity Absence and Uniformity Breach (e.g., a misaligned row of products, one tile laid upside-down) arise organically in our data but are rarely or never seen in synthetic datasets, which typically only show alterations of a single entity (see Figure \ref{fig:benchmark_comparison} for examples; taxonomies of datasets such as ROME \cite{zhou-etal-2023-rome} and GlitchBench \cite{taesiri2024glitchbench} only match with our Entity Attribute and Spatial Relation categories). Textual Anomalies are also usually absent from synthetic datasets, except MMIR \cite{yan2025multimodal}.

\textbf{Varied and Nuanced Anomalies:} The richness of CAVE is not just that real images are noisier; rather, the types of anomalies themselves are more varied and nuanced. Synthetic datasets are constrained by their generation rules: take an existing object or scene, alter it following a pre-defined template (e.g., change an object's color, swap animal species) to obtain an anomalous version. Given the limited set of alteration types, the resulting datasets are inherently limited in terms of anomaly diversity. In contrast, real data exposes new failure modes and contextual subtleties that template-driven methods cannot anticipate.

\subsection{Evaluation Methods}
Across these benchmarks, evaluation typically relies on zero-shot testing on large-scale pretrained models to assess how well they generalize to rare or absurd scenarios without task-specific adaptation. A few studies, like WHOOPS and HaloQuest, also explore fine-tuning on a training subset to boost performance, illustrating how effectively VLMs adapt to OOD data. In our study, we focus exclusively on zero-shot evaluation, as most anomalies in \ourdata are relatively easy for humans to identify (\Cref{fig:cave_stats_combined} (right)), and the small size of our dataset makes fine-tuning impractical.

\begin{table*}[h!]
    \centering
    \renewcommand{\arraystretch}{1.2} 
    \setlength{\tabcolsep}{6pt} 
    \small 
    \resizebox{\textwidth}{!}{
    \begin{tabular}{l cc cc c cccc}
        \toprule
        \textbf{Dataset} & \multicolumn{2}{c}{\textbf{Anomaly Type}} & \multicolumn{2}{c}{\textbf{Dataset Size}} & \textbf{Data source} & \multicolumn{3}{c}{\textbf{Task}} \\
        \cmidrule(r){2-3} \cmidrule(r){4-5} \cmidrule(r){7-10}
        & \textbf{Real} & \textbf{Synthetic} & \textbf{\#features} & \textbf{\#Images} & & \textbf{\#Anomaly tasks} & \textbf{Y/N} & \textbf{multi} & \textbf{Open} \\  
        \midrule

        Visual Riddles  & & \checkmark & 2 & 400 & Text-to-Image models & 1 &  & \checkmark & \checkmark \\  
        \addlinespace

        WHOOPS  & & \checkmark & 4 & 500 & Text-to-Image models &1 & & \checkmark & \checkmark  \\  
        \addlinespace

        HaloQuest  &  & \checkmark & 3 & 3,157 & Text-to-Image models + Open Images dataset & 1 & & & \checkmark \\  
        \addlinespace

        ROME  &  & \checkmark & 1 & 1,563 & ViComTe + ThingsNotWritten & 1& \checkmark & & \\  
        \addlinespace
        
        NEMO  &  & \checkmark & 1 & 900 & Text-to-image models & 1 & & \checkmark & \checkmark \\  
        \addlinespace
        
        ISEKAI  & & \checkmark & 1 & 1,498 & Text-to-Image models & 1 & & & \checkmark \\  
        \addlinespace
        
        PittAds & & \checkmark & 1 & 3,928 & Product ads \& public service announcements  & 3 &  & \checkmark \\  
        \addlinespace

        MMIR & & \checkmark & 1 & 534 & VisualWebArena, Zenodo & 2 & & \checkmark & \checkmark \\
        \addlinespace
        
        GlitchBench &  & \checkmark & 1 & 593 & Game-Physics dataset + Unity + YouTube & 1 & & & \checkmark \\  
        \addlinespace

        \midrule

        \ourdata{} & \checkmark & & 7 & 361 & Reddit & 3 & & \checkmark & \checkmark  \\  
         
        \bottomrule
    \end{tabular}}
    \caption{\textbf{Related Benchmarks.} Overview of multimodal reasoning benchmarks in images. Each benchmark is categorized based on the type of images it contains (real or synthetic), dataset scale (features per image and total number of images), generation method, and task involved (number of tasks related to anomaly, binary yes/no questions, multiple-choice VQA, and open-ended VQA).}
    \label{tab:benchmark_comparison}
\end{table*}

\begin{figure*}
    \centering
    \includegraphics[width=\linewidth]{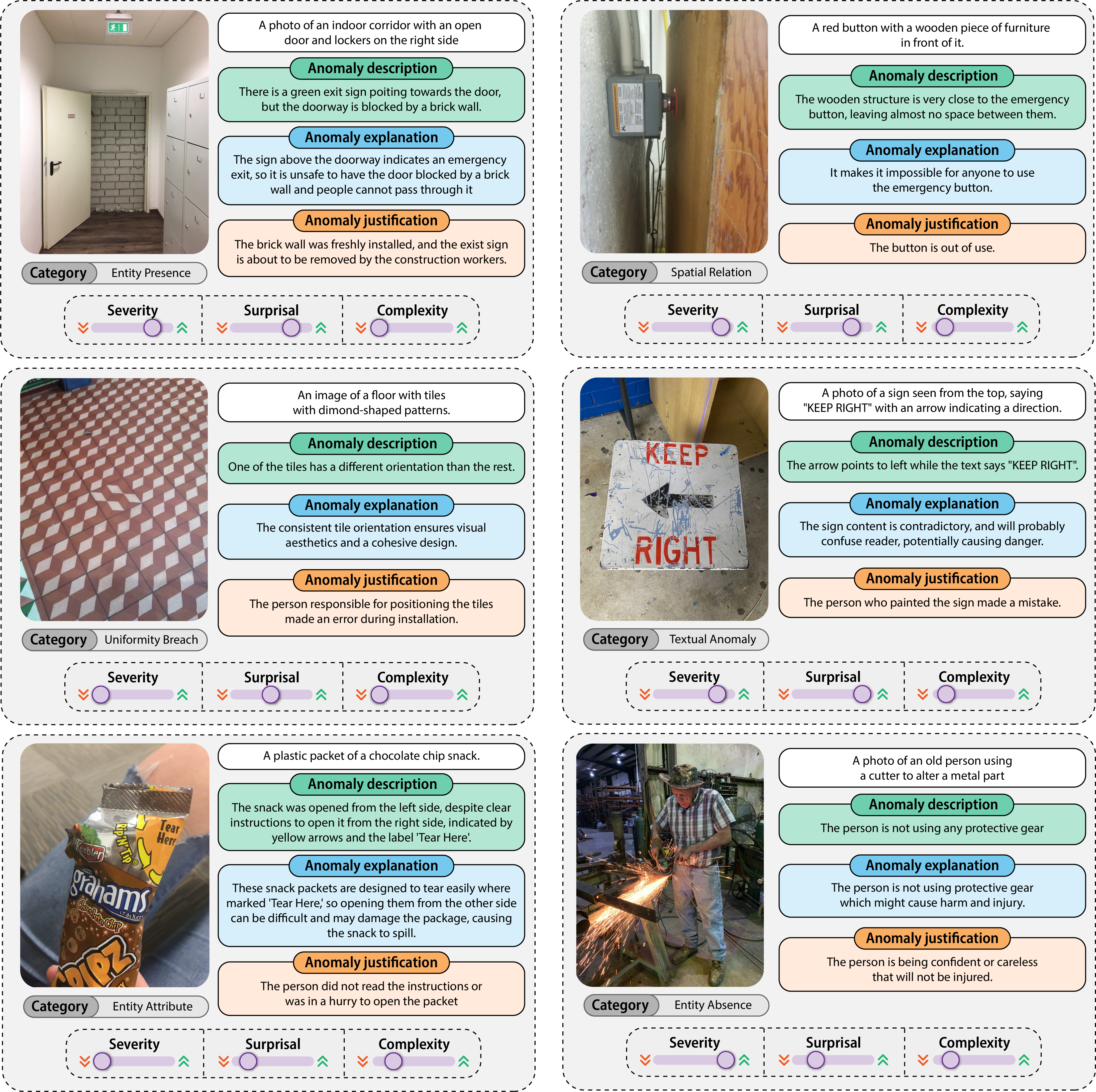}
    \caption{\textbf{Examples from \ourdata{}.} Each image is accompanied by a human-provided image description, anomaly description, anomaly explanation, anomaly justification, anomaly manifestation category, and numerical features of severity, surprisal, and complexity scores, for each of the anomaly manifestation categories present in CAVE.}
    \label{fig:examples-for-each-category}
\end{figure*}

\section{Quantitative Comparison with Synthetic Benchmarks}
 We perform a direct comparison with two prior synthetic benchmarks: WHOOPS, which uses full-image generation to create anomalous scenes, and COCO-OOC, which introduces anomalies via  image editing.

First, we evaluate WHOOPS, a benchmark closely aligned with our task formulation and commonsense-reasoning orientation. It provides ground truth anomaly descriptions, enabling the use of our full LLM-as-a-judge pipeline. Second, we sample 500 randomly selected anomalous images from COCO-OOC. Unlike WHOOPS, COCO-OOC modifies real COCO images by inserting localized anomalies while keeping the rest of the scene unaltered. However, the edits can introduce visual artifacts or noise that may affect anomaly detection. Since COCO-OOC lacks ground truth AD descriptions, we rely on manual evaluation to assess the correctness of the model-generated anomalies.

Across both benchmarks, we use identical evaluation prompts (GPT-4o with our AD prompt). We observe a substantial performance gap between CAVE and prior benchmarks as shown in \Cref{tab:synthetoic_benchmark_comparison}.

\begin{table}[h!]
\centering
\begin{tabular}{@{}lllr@{}}
\toprule
\textbf{Dataset} & \textbf{Precision} & \textbf{Recall} & \textbf{{F1-score}} \\
\midrule
WHOOPS & 85.5 & 85 & 85.3 \\
COCO-OOC & -  & 91 & -  \\
\textbf{CAVE} & \textbf{52.4} & \textbf{50} & \textbf{51.2}\\
\bottomrule
\end{tabular}
\caption{\textbf{Comparison with Synthetic Benchmarks.} We compare GPT-4o evaluation on WHOOPS and 500 samples of COCO-OOC. Note that COCO-OOC only has positive samples, hence we only report recall.}
\label{tab:synthetoic_benchmark_comparison}
\end{table}

 This demonstrates that CAVE is significantly more challenging for state-of-the-art VLMs. The higher precision on WHOOPS reflects its simpler, single-object or uncluttered scenes, which reduce hallucinations; in contrast, CAVE’s complex, real-world images introduce more distractors and plausible but subtle anomalies.

A manual analysis of false positives further underscores the difference: on WHOOPS, most FPs are perception errors, with only 4 reasoning errors. On CAVE, by contrast, reasoning errors account for about half of all FPs, indicating a much greater demand for contextual and commonsense reasoning (see \Cref{tab:count_error_analysis} and \Cref{fig:ex1} to \ref{fig:ex6}). Moreover, many of the WHOOPS FPs stem from artifacts of synthetic image generation (e.g., non-existent script, visual inconsistencies and missing elements). This leads to additional “valid” anomalies found by the model being marked as FPs, underestimating the model's performance.

We observed that the synthetic editing process (like in COCO-OOC) often results in unrealistic artifacts (e.g., partially added objects, such as a half-elephant). While this makes anomaly detection easier, it also limits ecological validity: models may succeed by flagging obvious artifacts rather than demonstrating true commonsense reasoning.

\section{Advanced Prompting Strategies}

\textbf{(1) Chain-of-thought prompting (CoT)} This strategy works by instructing models to ``think step by step" before answering, breaking complex reasoning into explicit sequential steps \cite{Wei2022ChainOT}. See prompt in \Cref{fig:AD_cot_prompt}.

\textbf{(2) Set-of-Marks prompting (SoM)}
We incorporate object-level annotations and bounding boxes generated by Grounding DINO \cite{Liu2023GroundingDM} to supplement the prompt with visual cues. Specifically, Grounding DINO identifies relevant regions in the image and provides precise bounding box coordinates, which serve as explicit visual references to guide the model’s attention. Each bounding box is labeled with a number in the top-left corner, indicating the detected object. Following \citet{yang2023setofmarkpromptingunleashesextraordinary}, we keep the textual prompt unchanged and instead replace the original images with versions that include these annotated boxes. As in the original work, the prompt does not explicitly mention the presence of bounding boxes.  This strategy aims to reduce perceptual errors, such as hallucinations or counting mistakes, by focusing the model’s attention on concrete visual entities \cite{yang2023setofmarkpromptingunleashesextraordinary}. The prompt used here is the vanilla inference prompt (see \Cref{fig:AD_prompt}).

\textbf{(3) Combined CoT+SoM prompting}
This strategy integrates the step-by-step reasoning of CoT with visual cues of SoM. This hybrid approach first establishes precise visual references using bounding boxes, then builds logical reasoning chains based on these grounded elements, enabling both spatial understanding and logical inference. The prompt used is identical to the CoT inference prompt (see \Cref{fig:AD_cot_prompt}), with the only change being the replacement of original images with versions containing bounding boxes.

\textbf{(4) Multi-step CoT prompting} 
Unlike standard CoT, this method decomposes the task into three sub-steps: (i) planning the reasoning process, (ii) identifying key visual elements, and (iii) generating anomaly descriptions based on these observations. Each sub-task is explicitly prompted, encouraging more organized and interpretable reasoning \cite{xu2025llavacotletvisionlanguage}. See prompt in \Cref{fig:AD_msr_prompt}.

\textbf{(5) CoT + Self-consistency prompting} , 
 In this strategy, the model is prompted multiple times (\textit{e.g.}, three) using the CoT format with stochastic sampling (temperature = 0.5). The resulting outputs are then aggregated using a majority-vote mechanism: only anomalies mentioned in at least two of the three generations are retained. This technique reduces spurious detections by encouraging agreement across multiple reasoning paths, effectively filtering out unstable or hallucinated outputs \cite{Wang2022SelfConsistencyIC}. See prompt in \Cref{fig:AD_self_consistency_prompt}.

\section{Human Annotations}\label{supp:data_annotation}

\subsection{Data Collection and Filtering}\label{supp:data_collection_filtering}

We scraped images from Reddit, focusing on four subreddits: \texttt{r/ocdtriggers}, \texttt{r/mildlyconfusing}, \texttt{r/mildlyinfuriating}, and \texttt{r/OSHA}. 
Using the PRAW\footnote{\url{https://github.com/praw-dev/praw}} library, we downloaded the top 1,000 posts from each subreddit. 
We kept only posts that contained images, and performed a first automated filtering, keeping only images above icon size.

We then manually filtered the remaining 1,725 images using the following criteria:
\begin{itemize}
    \item Remove toxic, harmful, and not safe for work content.
    \item Remove image featuring unrealistic content.
    \item Remove images with annotations: text added on top of the image, circles, etc. When possible, we manually edited images that could be cropped to hide the annotations on the image.
    \item Remove images that are ambiguous or have unidentifiable content.
\end{itemize}

Many samples contain anomalies that were done on purpose; often for convenience, but sometimes as a joke. We keep these ones, as detecting the presence of a visual anomaly created on purpose for humoristic purposes, and understanding why it is anomalous, is part of the VLM abilities we want to probe.


\subsection{Annotation round 1: Amazon Mechanical Turk}\label{supp:annotations_mturk}

We used Amazon Mechanical Turk to obtain annotations for the Reddit images. 
To ensure high-quality annotations, we conducted a worker selection round, ultimately selecting 40 workers for the task. Workers were pre-screened using Amazon Mechanical Turk’s automatic metrics with the following criteria: (a) HIT approval rate above 80\%, (b) location in the United States, and (c) more than 1,000 approved HITs.
Workers were compensated at a rate of 10 USD per hour, during the qualification and the annotation round. Each image received five annotations. We split the annotation into 3 rounds, allowing us to review the annotations between each round and provide feedback to the annotators when needed.

Below are the detailed instructions that were given to the annotators.

\begin{tcolorbox}[mysummarybox]





We need your help to identify and annotate anomalies in images. 
\textbf{An anomaly refers to anything that deviates from what most people consider standard, normal, or expected}. 
It can be an unusual element, action, or occurrence in an image that most people would find surprising or out of place.
For example, bowls of soup accompanied by forks but no spoon would be considered an anomaly because a spoon is expected for eating soup. In contrast, a plant placed on a computer desk is not an anomaly, as most people wouldn’t find it unusual.

\medskip

\textbf{Task Instructions:}  

\begin{enumerate}
    \item \textbf{Presence of Anomaly:}  Observe carefully the given image. Is there any anomalous element, according to the definition given above? Not all of the images necessarily have anomalies!  You can right-click on the images and select \textit{``Open in a new tab"} to zoom in.

    \item \textbf{Description of Anomaly:} Describe the image and the anomaly in detail:  What does the image show? What is abnormal or unexpected about it? Why is it considered an anomaly?

    
    \item \textbf{Type of Anomaly:}  
    \textbf{Select the type of anomaly} (an example for each type is given below):
    \begin{itemize}[leftmargin=0pt,labelindent=0pt]
        \item \textbf{Entity Presence:} Something is present in the image but shouldn’t be there.
        \item \textbf{Entity Absence:} Something that should be present is missing.
        \item \textbf{Entity Attribute:} An object has an anomalous attribute such as \textit{color, shape, label, orientation, or usage}.
        \item \textbf{Spatial Relation:} Something is incorrectly located or oriented relative to another element.
        \item \textbf{Uniformity Breach:} There is an unexpected or misplaced element in an ensemble that should be uniform or symmetrical.
        \item \textbf{Textual Anomaly:} The text in the image presents an unexpected, surprising, or illogical message.
    \end{itemize}
    \textbf{You may choose more than one type of anomaly if applicable.}
\end{enumerate}
\end{tcolorbox}

\subsection{Annotation round 2: Expert annotation consolidation}\label{supp:annotations_expert}

Following the first round, we manually filtered out samples that were confusing for annotators. Our pool of expert annotators includes undergraduate degree holders, graduate students, and PhD students with a background in NLP.

Below are the detailed instructions that were given to the annotators.  


\begin{tcolorbox}[mysummarybox]

{\large \textbf{Overview.}}


We are studying how well large vision-language models can identify anomalies that defy commonsense in images. Our goal is to assess their understanding of a situation, its severity, and potential solutions.

You will annotate anomalies visible in images. Each annotation form contains \textbf{5 images}. Each image has already been annotated by \textbf{4 to 5 workers} via MTurk, who answered the following questions:

\begin{enumerate}
    \item \textbf{Is there an anomaly in this image?}  
    \item \textbf{If yes, they described:}
    \begin{enumerate}
        \item \textbf{Anomaly Description (AD):} Describe the image and the anomaly in detail: what does the image show, what is wrong about it, and why?
        \item \textbf{Correct Version Description (CVD):} Describe what the correct version of the image would look like if the anomaly weren’t present.
    \end{enumerate}
\end{enumerate}

\textbf{Definition of an Anomaly}
An anomaly is anything that \textbf{deviates from what most people consider standard, normal, or expected}. It can be an unusual element, action, or occurrence in an image that would seem surprising or out of place to most people.

\textbf{Examples:}
\begin{itemize}
    \item A \textbf{bowl of soup served with a fork but no spoon} is an anomaly because a spoon is the expected utensil for soup.
    \item A \textbf{plant on a computer desk} is \textbf{not} an anomaly, as it is a common and expected item in such a setting.
\end{itemize}

\textbf{Key Principle:} Identifying an anomaly should rely \textbf{only on what is clearly visible in the image}—it should not require excessive assumptions about the situation.

\textbf{Don’t spend too much time on a single image.} 
If you’re unsure or confused about an image or an annotation, \textbf{skip it} and leave a note in the open field at the bottom of the page.

\medskip

{\large \textbf{Instructions.}}

Workers often identified different anomalies in the same image. Your task is to consolidate their annotations into a structured format. You may input \textbf{up to 3 anomalies} per image. Most images contain only one anomaly. For each image, based on the workers' annotations, provide a final set of anomalies in the following format:

\begin{enumerate}[leftmargin=9pt,labelindent=0pt]
    \item \textbf{Image Description}: Provide a short description of the image, without describing the anomaly. Include any useful context, such as whether the image is a \textit{photo, screenshot, or illustration}, the \textit{location}, etc.
    \item \textbf{Anomaly Description (AD)}: Clearly describe the anomaly.
    \item \textbf{Correct Version Description (CVD)}: Describe what the image would look like if the anomaly weren’t present. \textbf{Do not} describe how to fix the anomaly—only describe the correct version as if it were normal.
    \item \textbf{Anomaly Explanation (AE)}: Explain why it is anomalous. Avoid vague statements like “because it’s abnormal.” Instead, consider: \textit{Why is the correct version expected? What makes the anomaly logically inconsistent or unexpected?}
    \item \textbf{Anomaly Justification}: Provide a realistic and plausible explanation for how the anomaly might have occurred. Keep it concise (\textbf{max 2 sentences}).  
    Example: If an object is blocking a door, a plausible justification might be: \textit{“The door is not in use because it leads to an empty space.”}
    \item \textbf{Anomaly Severity (Does the anomaly require immediate action?)}

\begin{itemize}
    \item \textbf{1} = Does not require action; purely aesthetic or has no impact on functionality/safety.  
    \textit{Example: A small stain on a non-critical surface.}
    \item \textbf{3} = Moderately concerning; might cause inconvenience or minor inefficiencies but does not pose immediate risks.  
    \textit{Example: A misaligned sign that is still readable.}
    \item \textbf{5} = Requires immediate action; it could present a safety hazard, major malfunction, or significant risk.  
    \textit{Example: A worker using a circular saw without protection gear.}
\end{itemize}

     \item \textbf{Anomaly Surprisal (How much does it deviate from expectations?)}
    
\begin{itemize}
    \item \textbf{1} = Common, not very surprising; frequently observed in similar contexts.  
    \textit{Example: A car parked in an inconvenient way.}
    \item \textbf{3} = Unusual but not shocking; uncommon but plausible.
    \item \textbf{5} = Extremely rare and highly surprising; would cause strong reactions (shock, confusion, amazement).  
    \textit{Example: A tree growing upside down from a roof.}
\end{itemize}

    \item \textbf{Anomaly Complexity (How hard was this anomaly to detect?)}
    
\begin{itemize}
    \item \textbf{1} = Obvious and easy to notice; immediately stands out.  
    \textit{Example: A red apple in a pile of green apples.}
    \item \textbf{3} = Requires some attention to notice; not the first thing seen but becomes clear after a few seconds.  
    \textit{Example: A misspelled word on a sign.}
    \item \textbf{5} = Very hard to detect; blends into the environment or requires specific knowledge to identify.  
    \textit{Example: A minor defect in complex machinery.}
\end{itemize}

\end{enumerate}

{\large \textbf{Guidelines:}}  

\medskip

In practice, you will reuse the MTurk annotations. Here are common situations you may encounter and how to handle them:

\begin{itemize}[leftmargin=7pt,labelindent=0pt]
    \item \textbf{Same Anomaly from Different Workers.}  
    If multiple workers describe the same anomaly, \textbf{merge their descriptions} into one clear and accurate version.  
    \textbf{Two anomalies are the same if they have the same description and explanation.}

    \item \textbf{One Worker Describes Multiple Anomalies Jointly.}  
    If a worker describes multiple anomalies together, \textbf{split them into separate entries} and fill in the necessary fields for each.

    \item \textbf{Invalid Anomaly.}  
    \begin{itemize}
        \item Does this truly qualify as an anomaly based on the definition?
        \item Did the worker make \textbf{assumptions} about the situation that are not straightforward using the image alone?
        \item Did the worker \textbf{misinterpret} the image?
    \end{itemize}
    If invalid, \textbf{flag it and do not include it} in the consolidated list.

    \item \textbf{Unclear Anomaly Description.}  
    If an anomaly is valid but \textbf{poorly described}, \textbf{rephrase it clearly} and complete the required fields (AD, AE, CVD, etc.).

    \item \textbf{Unclear or Incorrect Correct Version Description (CVD).}  
    If a worker's CVD does not align with the anomaly or is poorly phrased, \textbf{rewrite it according to the guidelines}.

    \item \textbf{No Workers Found an Anomaly.}  
    If no worker identified an anomaly, \textbf{check if you can spot an obvious one}.  
    If not, \textbf{leave the fields empty}.

    \item \textbf{All Reported Anomalies Are Invalid.}  
    If none of the workers' anomalies match the definition and you don’t see any other valid anomaly, \textbf{leave everything empty}.
\end{itemize}

{\large \textbf{In practice:}}  

\textbf{For convenience, you can:}  
    \begin{itemize}
        \item Copy-paste the list of MTurk annotations to the side for easy reference.
        \item Open the image in full resolution in another window.
        \item Keep these instructions open in a separate tab.
    \end{itemize}

\textbf{LLM Usage:}  
    \begin{itemize}
        \item You can use a language model to check and correct the grammar of your annotations.
        \item \textbf{DO NOT upload or share the image with an LLM!}
    \end{itemize}

\end{tcolorbox}

\subsection{Numerical features inter-rater agreement}\label{supp:annotator_agreement}

Each numerical feature -- anomaly surprisal, complexity and relevance -- is annotated by 3 people. We measure the agreement between the 3 annotators (\cref{tab:numerical_feature_agreement}) using Spearman’s Rank Correlation, Krippendorff’s Alpha, and Gwet’s AC2.  
Spearman’s Rank Correlation (0.65) and Krippendorff’s Alpha (0.62) indicate moderate-to-strong agreement among annotators for severity, and weaker for surprisal, which is more sujective. Since surprisal and complexity are imbalanced, we turn to Gwet's AC2 \cite{gwet2008computing}, a paradox-resistant agreement score, where the chance agreement is measured in a less distribution-sensitive fashion. We use quadratic weights, meaning that larger disagreements are exponentially more problematic than smaller ones. Indeed, likert-scale ratings with relatively subjective tasks such as here may lead to confusions between similar ratings (4 and 5, 1 and 2). Gwet's AC2 highlights a much higher agreement for the complexity score of 0.76, which is considered good \cite{gwet2014handbook}.

\begin{table}[!ht]
    \centering
    \footnotesize
 \resizebox{\columnwidth}{!}{%
\begin{tabular}{lrrr}
\toprule
{} & \textbf{Spearman $\rho$} &\textbf{ Krippendorff $\alpha$}& \textbf{Gwet AC2} \\
\midrule
\textbf{Severity}   & 0.65 & 0.62 & 0.58 \\
\textbf{Surprisal}  & 0.34 & 0.32 & 0.54 \\
\textbf{Complexity} & 0.28 & 0.23 & 0.76 \\
\bottomrule
\end{tabular}
}
    \caption{Inter-rater agreement for each numerical feature.}
    \label{tab:numerical_feature_agreement}
\end{table}

\subsection{Cultural representation \& bias}
\label{supp:cultural_bias}

\begin{figure*}
    \centering
    \includegraphics[width=\linewidth]{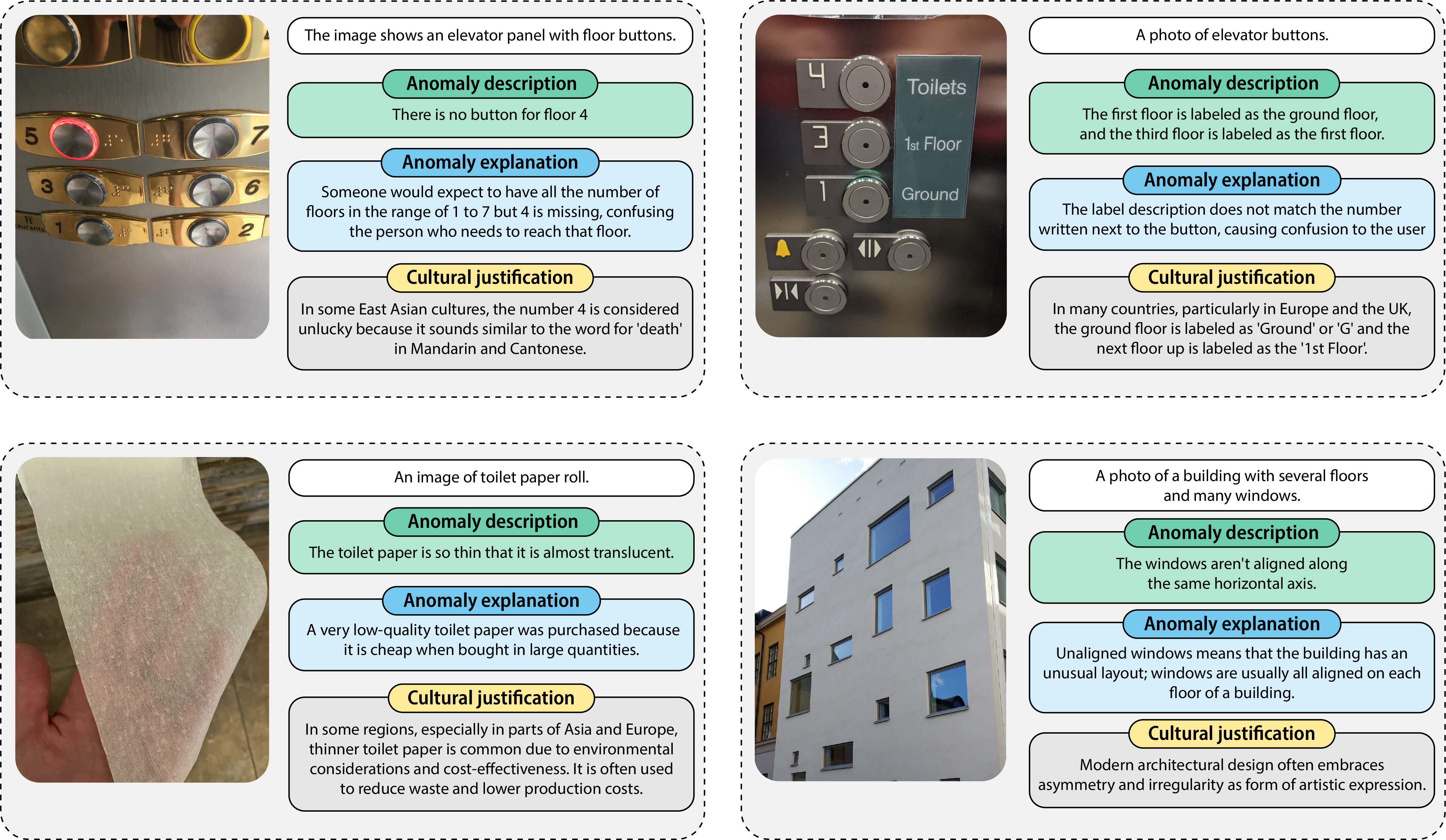}
    \caption{\textbf{Culture-specific examples of CAVE.} Examples of anomalies from \ourdata{} annotated as Western-centric, along with culturally grounded justifications explaining why they should not be considered anomalies.}
    \label{fig:cultural-biased}
\end{figure*}

An anomaly is generally defined as a deviation from the norm. In this context, "norm" refers to a set of expectations commonly shared within a particular social or cultural group. Some of these norms are broadly universal, for example, adhering to safety standards to avoid hazardous situations, while others are culturally specific, such as the custom of wearing red at weddings in China \cite{goto2010cultural, myung2024blend, nayak2024benchmarking}. As a result, interpretations of what constitutes an anomaly can differ significantly across cultural contexts, leading to situations that may appear ordinary to individuals from one background and anomalous to those from another \cite{ye2023computer}.

To explore the extent to which cultural bias influences the perception of anomalies, we conducted an analysis of the CAVE dataset. Specifically, we examined whether a subset of visual anomalies presented in the dataset reflected culturally contingent interpretations. 
We selected a subset of 35 anomalies based on high variance (above 1.5 for each feature) in the numerical features obtained from annotator responses, as this variance suggests a lack of consensus that may be attributable to differing cultural perspectives.
Among these, we identified four images containing anomalies that appeared Western-centric but would not be considered anomalous in other cultural contexts. In addition, from the full benchmark, we selected 20 examples reflecting personal biases, such as anomalies related to how individuals park their cars or behave in public spaces, as well as a set of universally recognized anomalies. For each of these 24 samples, we provided explanations of the relevant cultural context, where applicable, and updated the corresponding Anomaly Justification (AJ) annotations accordingly.
Using these manually curated annotations as reference labels, we constructed a prompt to evaluate whether each anomaly aligned with specific cultural, religious, regional, or historical norms, and not with personal biases. This prompt was submitted to GPT-4o for analysis on the same subset. The model performed well, misclassifying only one instance: a train seat colored differently from the rest. While this was intended to reflect a "uniformity breach," the model interpreted it as a designated priority seat—an error likely due to contextual ambiguity.

We subsequently applied the same automatic bias assessment method to the entire CAVE dataset to verify the initial manual annotation. This broader analysis identified the same four anomalies that exhibited a Western-centric bias. These instances are presented in Figure \ref{fig:cultural-biased}, along with the model’s culturally influenced anomaly justifications for each.
This analysis indicates that while the majority of anomalies in the CAVE dataset are perceived as universally anomalous and actionable, a small number are influenced by culturally specific norms, particularly those aligned with Western perspectives. These findings underscore the importance of accounting for cultural variability in the development of robust and inclusive anomaly detection systems.

\section{Prompts}\label{supp:prompts}
The prompts for six tasks, the automatic evaluation and the cultural assessment are listed below:
\begin{itemize}
    \item Anomaly Description: \Cref{fig:AD_prompt}
    \item Anomaly Explanation: \Cref{fig:AE_prompt}
    \item Anomaly Justification: \Cref{fig:AJ_prompt}
    \item Anomaly Severity: \Cref{fig:severity_prompt}
    \item Anomaly Surprisal: \Cref{fig:sup_prompt}
    \item Anomaly Complexity: \Cref{fig:comp_prompt}
    \item AD judge prompt: \Cref{fig:ad_judge_prompt}
    \item AE judge prompt: \Cref{fig:ae_judge_prompt}
    \item Cultural bias assessment prompt: \Cref{fig:cultural_bias_prompt}
\end{itemize} 

\section{Additional Results}

\subsection{Anomaly Description}\label{app:AD_results}

\begin{figure}[h!]
    \centering
    \includegraphics[width = \linewidth]{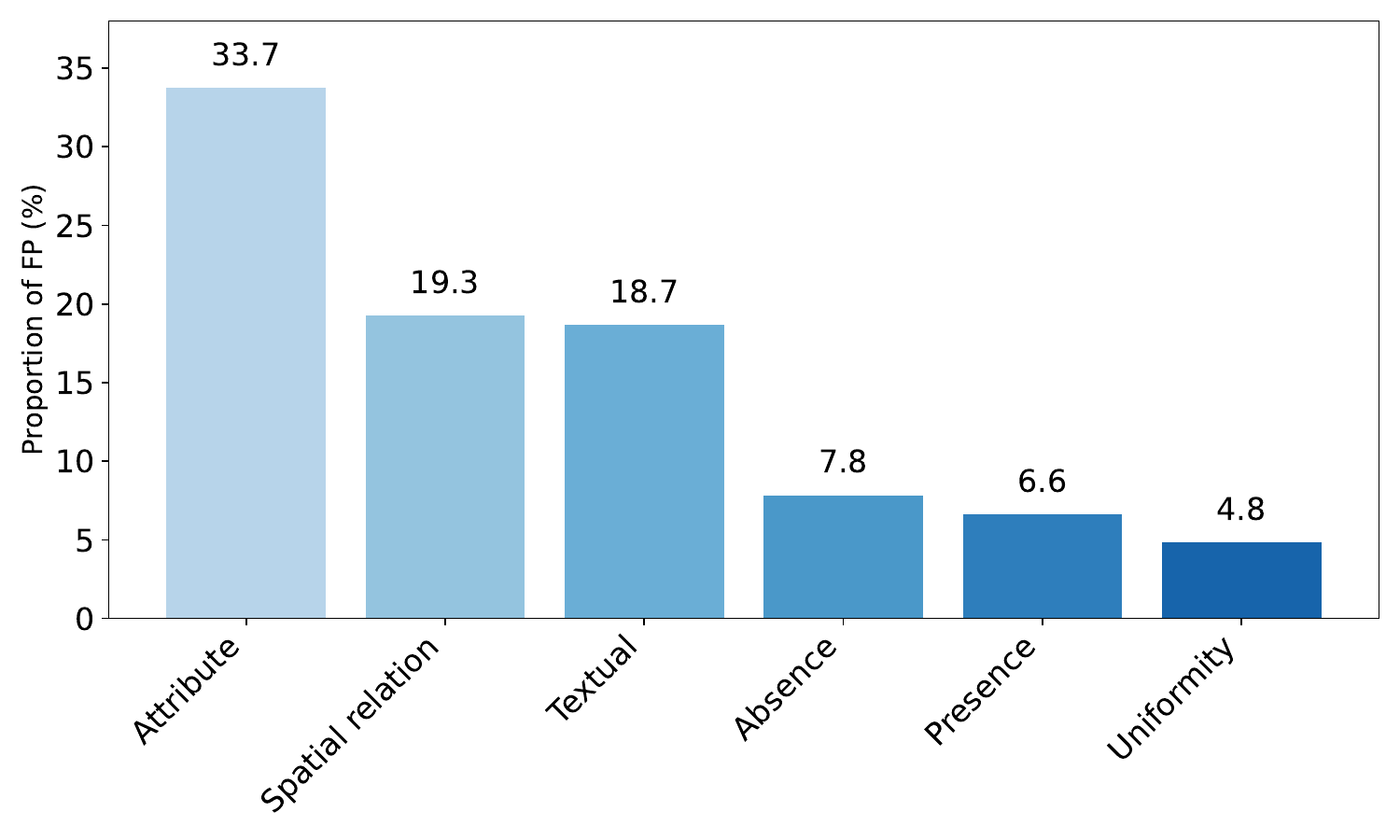}
\caption{\textbf{FP classifier performance.} Anomaly category proportions in GPT-4o FP.}
\label{fig:anomaly-fp}
\end{figure}

WE categorize all false positives (anomalies hallucinated by the VLM) into the different anomaly visual manifestation types (according to our taxonomy), by tuning a classifier of Anomaly Descriptions. We run the classifier on GPT-4o’s false positives using the prompt given in \Cref{fig:fp_classifier_prompt}. \Cref{fig:anomaly-fp} shows that GPT-4o predominantly hallucinates anomalous entity attributes (\textit{e.g.}, count, color), anomalous spatial relations, and textual anomalies (anomalies in the context of text seen in the image).

We also performed a bootstrap test on 5k different random samples of CAVE to assess significance for prompting strategies over the vanilla baseline. \Cref{tab:bootstrap_results} reports the improvement in F1 scores for the 3 best prompting strategies compared to the vanilla prompt baseline, with 95\% confidence intervals shown in brackets and corresponding $p$-values, in bold for statistically significant gains ($p$ \(\leq\) 0.05). 
Overall, CoT with self-consistency yields significant gains for most models, whereas plain CoT and multi-step CoT reach significance for roughly half.

\begin{table*}[!h]
\centering
\small
\begin{tabular}{@{}llll@{}}
\toprule
\textbf{Model} & \textbf{CoT} & \textbf{CoT + consist.} & \textbf{MS CoT} \\ \midrule
Llama3.2 90b & 11.145 {[}0.059, 0.160{]} \textit{\textbf{p=0.000}} & 13.553 {[}0.081, 0.188{]} \textit{\textbf{p=0.000}} & 7.160 {[}0.020, 0.120{]} \textit{\textbf{p=0.003}} \\
LlavaOV 72b & -0.160 {[}-0.043, 0.040{]} \textit{p=0.524} & 8.780 {[}0.042, 0.133{]} \textit{\textbf{p=0.000}} & 2.088 {[}-0.028, 0.067{]} \textit{p=0.204} \\
InternVL2.5 38b & 2.937 {[}-0.006, 0.065{]} \textit{p=0.056} & 6.284 {[}0.022, 0.103{]} \textit{\textbf{p=0.002}} & -1.330 {[}-0.054, 0.027{]} \textit{p=0.732} \\

QwenVL2.5 72b & -2.845 {[}-0.068, 0.010{]} \textit{p=0.925} & -1.468 {[}-0.053, 0.024{]} \textit{p=0.764} & -1.609 {[}-0.057, 0.025{]} \textit{p=0.773} \\
InternVL2.5 78b & 2.411 {[}-0.018, 0.066{]} \textit{p=0.125} & 3.207 {[}-0.005, 0.069{]} \textit{\textbf{p=0.044}} & -0.876 {[}-0.049, 0.031{]} \textit{p=0.659} \\
\midrule
GPT-4o & 3.113 {[}-0.006, 0.068{]} \textit{\textbf{p=0.049}} & 2.543 {[}-0.010, 0.061{]} \textit{p=0.081} & 4.954 {[}0.010, 0.089{]} \textit{\textbf{p=0.007}} \\
Claude & 8.051 {[}0.026, 0.133{]} \textit{\textbf{p=0.001}} & 6.155 {[}0.015, 0.107{]} \textit{\textbf{p=0.006}} & 16.384 {[}0.120, 0.206{]} \textit{\textbf{p=0.000}} \\ \bottomrule
\end{tabular}
\caption{\textbf{Effectiveness of prompting strategies via bootstrapping.} Each entry shows the improvement over vanilla AD prompt, 95\% confidence interval, and $p$-value. Bold indicates statistically significant improvements ($p<0.05$).}
\label{tab:bootstrap_results}
\end{table*}

\subsection{Anomaly Explanation}\label{app:AE_results}
Each model's performance on TP and FN from the AD task is detailed in \Cref{tab:ae-tp-vs-fn-results}. Most of the models have higher performance on TP examples than FN.
\begin{table}[h!]
\centering
\small
\label{tab:ae-accuracy}
\begin{tabular}{lcc}
\toprule
\textbf{Model} & \textbf{TP Acc. (\%)} & \textbf{FN Acc. (\%)} \\
\midrule
\textit{open-source models} \\
Llama3.2 90b     & 82.22 & 76.88 \\
LlavaOV  72b        & 90.67 & 79.76 \\
InternVL2.5 38b      & 84.26 & 84.21 \\
QwenVL2.5 72b        & 87.39 & 82.64 \\
InternVL2.5 78b      & 81.08 & 86.58 \\
\midrule
\textit{closed-source models}\\
GPT-4o                  & 90.86 & 85.22 \\
o1                      & 93.02 & 88.89 \\
Claude & 87.10 & 73.97\\
\midrule
Average & 83.97 & 81.02 \\
\bottomrule
\end{tabular}
\caption{\textbf{AE Results on TP vs FN.} AE Accuracy on TP vs FN for each model.}
\label{tab:ae-tp-vs-fn-results}
\end{table}

\begin{figure*}
    \centering
    \includegraphics[width=\linewidth]{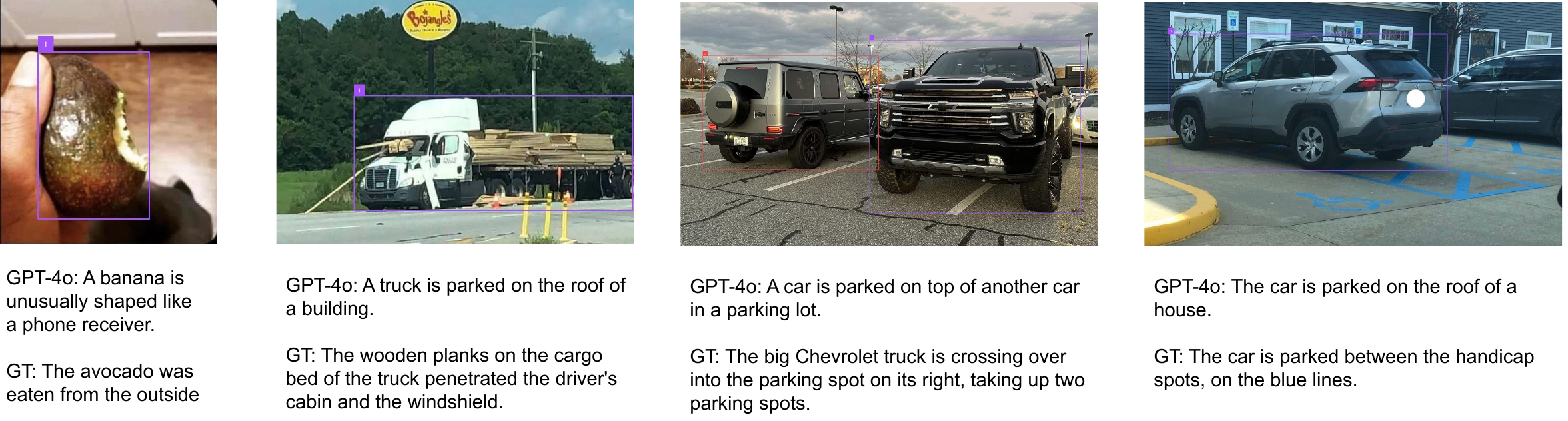}
    \caption{\textbf{Set-of-Marks images.} GPT-4o anomaly descriptions based on images with bounding boxes derived from GroundingDINO.}
    \label{fig:som_outputs}
\end{figure*}

\begin{table*}[ht!]
\centering
\small
\begin{tabular}{lcccccc}
\toprule
\textbf{Prompting Strategy} & \textbf{TP} & \textbf{FP} & \textbf{FN} & \textbf{Precision} & \textbf{Recall} & \textbf{F1 Score} \\
\midrule
Vanilla & 119.75 & 191.38 & 219.13 & 41.19 & 35.37 & 37.35 \\
CoT & 139.63 {\scriptsize\textcolor{darkgreen}{(+19.88)}} & 159.00 {\scriptsize\textcolor{darkgreen}{(-32.38)}} & 247.38 {\scriptsize\textcolor{red}{(+28.25)}} & 47.95 {\scriptsize\textcolor{darkgreen}{(+6.76)}} & 36.08 {\scriptsize\textcolor{darkgreen}{(+0.71)}} & 40.90 {\scriptsize\textcolor{darkgreen}{(+3.55)}} \\
SoM & 136.00 {\scriptsize\textcolor{darkgreen}{(16.25)}} & 222.38 {\scriptsize\textcolor{red}{(+31.00)}} & 251.00 {\scriptsize\textcolor{red}{(+31.88)}} & 43.38 {\scriptsize\textcolor{darkgreen}{(+2.20)}} & 35.14 {\scriptsize\textcolor{red}{(-0.23)}} & 37.35 {\scriptsize\textcolor{darkgreen}{(+0.00)}} \\
CoT+SoM & 123.50 {\scriptsize\textcolor{darkgreen}{(+3.75)}} & 181.13 {\scriptsize\textcolor{darkgreen}{(-10.25)}} & 263.50 {\scriptsize\textcolor{red}{(+44.38)}} & 42.01 {\scriptsize\textcolor{darkgreen}{(+0.83)}} & 31.91 {\scriptsize\textcolor{red}{(-3.46)}} & 35.91 {\scriptsize\textcolor{red}{(-1.44)}} \\
MS CoT & 144.50 {\scriptsize\textcolor{darkgreen}{(+24.75)}} & 150.13 {\scriptsize\textcolor{darkgreen}{(-41.25)}} & 242.50 {\scriptsize\textcolor{red}{(+23.38)}} & 50.45 {\scriptsize\textcolor{darkgreen}{(+9.26)}} & 33.88 {\scriptsize\textcolor{red}{(-1.49)}} & 40.18 {\scriptsize\textcolor{darkgreen}{(+2.82)}} \\
Self-consistency & 145.13 {\scriptsize\textcolor{darkgreen}{(+25.38)}} & 141.75 {\scriptsize\textcolor{darkgreen}{(-49.63)}} & 240.63 {\scriptsize\textcolor{red}{(+21.50)}} & 51.72 {\scriptsize\textcolor{darkgreen}{(+10.53)}} & 37.50 {\scriptsize\textcolor{darkgreen}{(+2.13)}} & 43.10 {\scriptsize\textcolor{darkgreen}{(+5.75)}} \\ \bottomrule
\end{tabular}
\caption{\textbf{Overall anomaly detection performance.} Values in parentheses indicate deltas from the Vanilla baseline; \textcolor{darkgreen}{green} with for improvement, \textcolor{red}{red} for decline.}
\label{tab:prompt_performance}
\end{table*}

\begin{table*}[ht!]
\centering
\small
\begin{tabular}{lccccccc}
\toprule
\textbf{Prompting Strategy} & \textbf{Absence} & \textbf{Attribute} & \textbf{Presence} & \textbf{Relation} & \textbf{Textual} & \textbf{Uniformity} \\
\midrule
Vanilla	&24.78	&35.10	&51.13	&32.02	&53.00	&28.86\\
CoT & 30.84 {\scriptsize\textcolor{darkgreen}{(+6.05)}} & 39.13 {\scriptsize\textcolor{darkgreen}{(+4.03)}} & 56.95 {\scriptsize\textcolor{darkgreen}{(+5.82)}} & 35.62 {\scriptsize\textcolor{darkgreen}{(+3.60)}} & 60.58 {\scriptsize\textcolor{darkgreen}{(+7.58)}} & 30.56 {\scriptsize\textcolor{darkgreen}{(+1.71)}} \\
SoM & 27.03 {\scriptsize\textcolor{darkgreen}{(+2.25)}} & 33.58 {\scriptsize\textcolor{red}{(-1.52)}} & 47.46 {\scriptsize\textcolor{red}{(-3.67)}} & 30.57 {\scriptsize\textcolor{red}{(-1.45)}} & 55.36 {\scriptsize\textcolor{darkgreen}{(+2.36)}} & 25.95 {\scriptsize\textcolor{red}{(-2.91)}} \\
SoM+CoT & 27.85 {\scriptsize\textcolor{darkgreen}{(+3.06)}} & 33.20 {\scriptsize\textcolor{red}{(-1.91)}} & 52.85 {\scriptsize\textcolor{darkgreen}{(+1.72)}} & 30.74 {\scriptsize\textcolor{red}{(-1.28)}} & 54.61 {\scriptsize\textcolor{darkgreen}{(+1.61)}} & 28.38 {\scriptsize\textcolor{red}{(-0.48)}} \\
MS CoT & 26.74 {\scriptsize\textcolor{darkgreen}{(+1.95)}} & 39.90 {\scriptsize\textcolor{darkgreen}{(+4.79)}} & 53.44 {\scriptsize\textcolor{darkgreen}{(+2.31)}} & 33.63 {\scriptsize\textcolor{darkgreen}{(+1.61)}} & 57.15 {\scriptsize\textcolor{darkgreen}{(+4.15)}} & 27.35 {\scriptsize\textcolor{red}{(-1.51)}} \\
Self-consistency & 31.97 {\scriptsize\textcolor{darkgreen}{(+7.19)}} & 41.83 {\scriptsize\textcolor{darkgreen}{(+6.73)}} & 56.12 {\scriptsize\textcolor{darkgreen}{(+4.99)}} & 36.52 {\scriptsize\textcolor{darkgreen}{(+4.50)}} & 56.97 {\scriptsize\textcolor{darkgreen}{(+3.97)}} & 32.45 {\scriptsize\textcolor{darkgreen}{(+3.59)}}\\
\midrule
\textbf{Average} &	28.20 &	37.12	& 52.99	& 33.18	& 56.28 & 28.92 \\
\midrule
\textbf{Rank} & 6 & 3 & 2 &4 & 1 & 5  \\
\bottomrule
\end{tabular}
\caption{\textbf{F1 scores per anomaly category.} Values in parentheses indicate deltas from the Vanilla baseline; \textcolor{darkgreen}{green} for improvement, \textcolor{red}{red} for decline.}
\label{tab:ad_per_category}
\end{table*}

\begin{figure}[!ht]
    \centering
    \includegraphics[width=\linewidth]{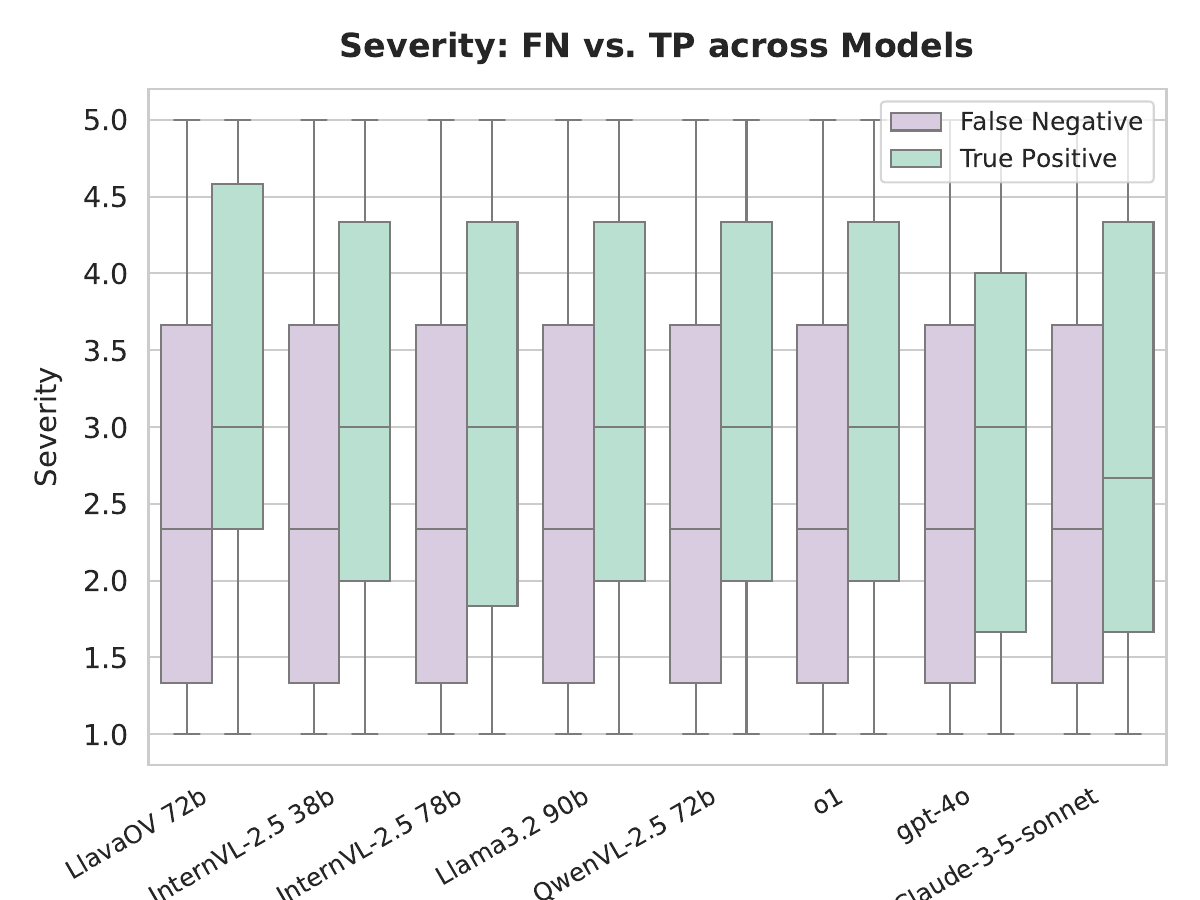}
    \caption{\textbf{Models' performance across severity feature.} Plot showing the deviation in the models' performance across different levels of anomaly severity for the anomaly description task. The results indicate that models perform well on less severe anomalies, while performance drops significantly for highly severe anomalies on average.}
    \label{fig:app-AD-severity}
\end{figure}
\begin{figure}[!ht]
    \centering
    \includegraphics[width=\linewidth]{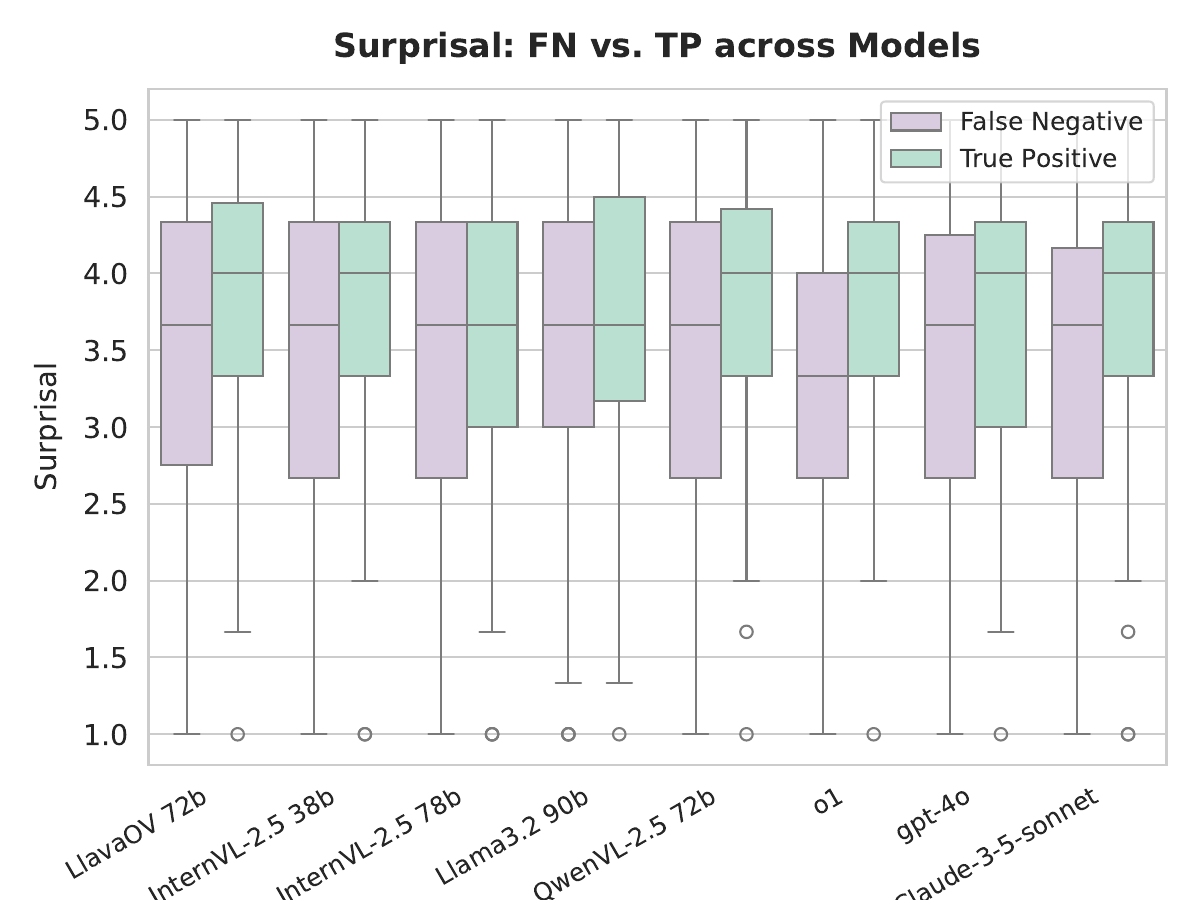}
    \caption{\textbf{Models' performance across surprisal feature.} Plot showing the deviation in the models' performance across different levels of anomaly surprisal for the anomaly description task. The results reveal that models perform well on high-surprisal anomalies but also exhibit more false positives for more surprising anomalies on average.}
    \label{fig:app-AD-surprise}
\end{figure}
\begin{figure}
    \centering
    \includegraphics[width=\linewidth]{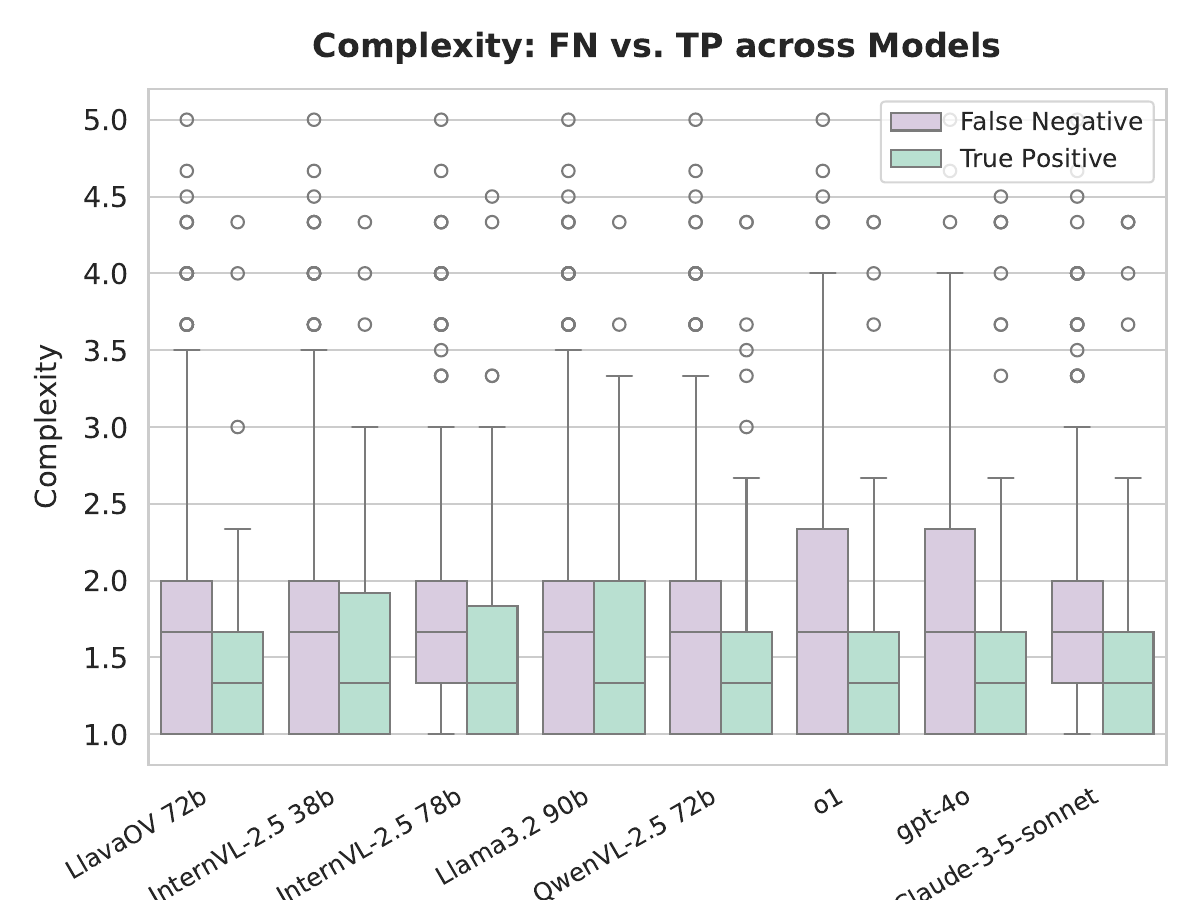}
    \caption{\textbf{Models' performance across complexity feature.} Plot showing the deviation in the models' performance across different levels of anomaly complexity for the anomaly description task. The results reveal that models perform well only on low-complex tasks but also exhibit false positives for much simpler anomalies on average.}
    \label{fig:app-AD-complexity}
\end{figure}

\subsection{Anomaly Justification}\label{app:AJ_results}
\begin{figure}[!ht]
    \centering
    \includegraphics[width=\linewidth]{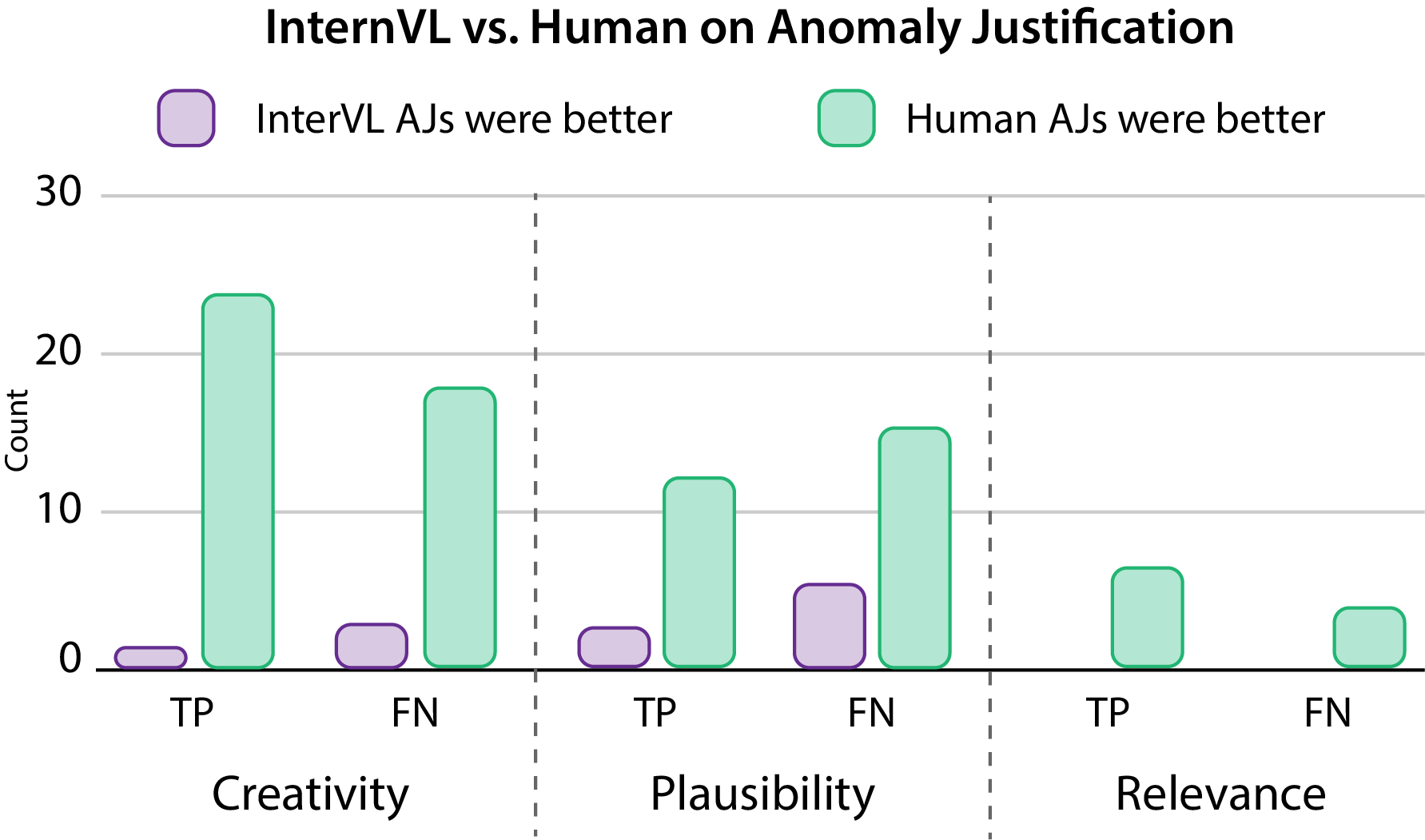}
    \caption{\textbf{Comparison of InternVL vs. Human Anomaly Justification.} Bars above the x-axis indicate cases where InternVL outperformed humans, while bars below indicate cases where InternVL underperformed. The 3 bars on the left are results over 50 False Negatives (FN), where the model failed to identify anomalies; the 3 bars on the right are over 50 True Positives (TP).}
    \label{fig:internVL_AJ_eval}
\end{figure}

\Cref{fig:internVL_AJ_eval} compares InternVL2.5 78B with human anomaly justifications.

\subsection{Numerical features prediction}\label{app:numerical_features_results}

The last set of tasks of \ourdata{} is the classification of the anomalies into ordinal features: surprisal, severity, and complexity, across a scale of 1 to 5. Echoing the inter-rater agreement that we computed between the 3 expert annotators on the surprisal, severity, and complexity scores, we measure the agreement between the human average score for each feature and the models' predictions of each score. The prompts used for these features can be found in \cref{supp:prompts}.

\begin{table}[!ht]
    \centering
    \begin{tabular}{lcc|cc}
        \toprule
        & \multicolumn{2}{c}{\textbf{GPT-4o}} & \multicolumn{2}{c}{\textbf{InternVL2.5 78b}} \\
        \cmidrule(lr){2-3} \cmidrule(lr){4-5}
        &  \textbf{$\rho$} & \textbf{AC2} & \textbf{ $\rho$} & \textbf{AC2} \\
        \midrule
        \textbf{Severity}   & 0.78 & 0.79 & 0.75 & 0.77 \\
        \textbf{Surprisal}  & 0.49 & 0.81 & 0.28 & 0.24 \\
        \textbf{Complexity} & 0.27 & 0.80 & 0.26 & 0.61 \\
        \bottomrule
    \end{tabular}
    \label{tab:numerical_feature_prediction}
        \caption{\textbf{Numerical Feature Prediction.} Comparison of GPT-4o and InternVL2.5 78b prediction of Anomaly Severity, Surprisal and Complexity. We measure Gwet's AC2 and Spearman's $\rho$.}
\end{table}


GPT-4o and InternVL show high agreement with humans for severity (\cref{tab:numerical_feature_prediction}), with both models achieving strong agreement scores. Surprisal and complexity prediction are harder tasks for both models.

\begin{figure*}[h!]
    \centering
    \includegraphics[width=\textwidth]{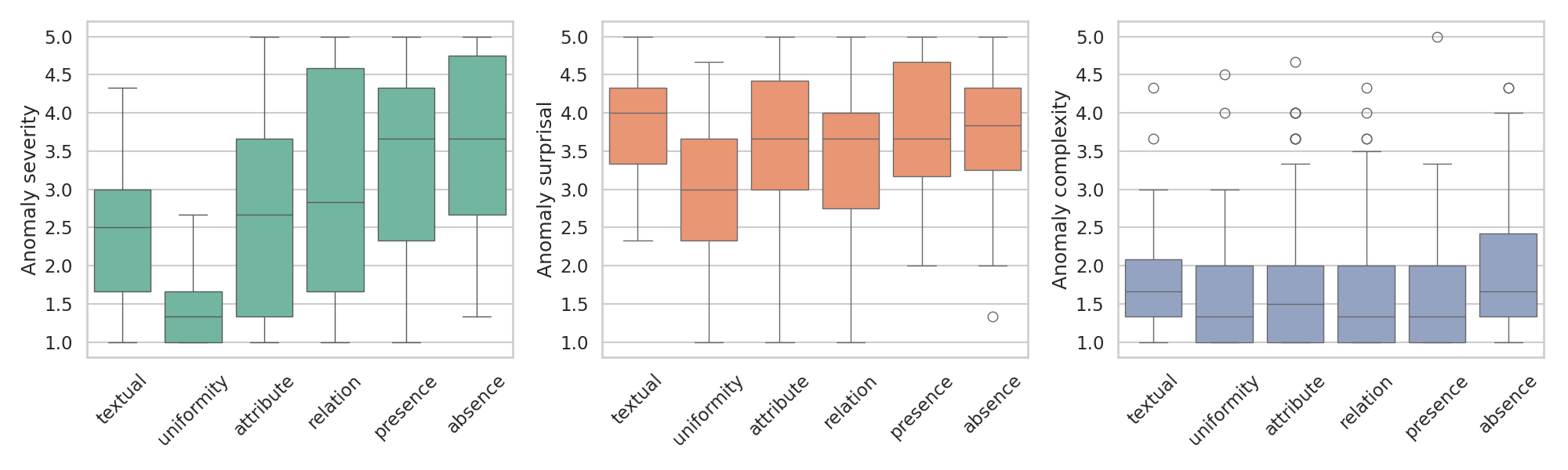}
    \caption{Distribution of anomaly scores across categories. The boxplots illustrate the distribution of complexity, severity, and surprisal scores across different anomaly categories, highlighting variations in human perception of anomalies.}
    \label{fig:category_per_feature}
\end{figure*}

The analysis of complexity, severity, and surprisal scores across different anomaly categories has been shown in \Cref{fig:category_per_feature}. The severity scores indicate anomalies categorized under entity absence and presence tend to be perceived as more severe. Conversely, anomalies related to uniformity breaches are consistently viewed as less severe. Examining the complexity scores, we observe that categories like textual anomalies exhibit greater variability, suggesting diverse perceptions of complexity within annotators, whereas uniformity anomalies show lower complexity scores with minimal variance. The distribution of surprisal scores indicates that anomalies in the textual and presence categories consistently evoke stronger feelings of unexpectedness, while again, anomalies categorized as uniformity remain at lower surprise levels.

\begin{table*}[h!]
\tiny
\centering
\resizebox{\textwidth}{!}{%
\begin{tabular}{@{}lcccccc@{}}
\toprule
\textbf{Model}  & \multicolumn{1}{l}{\textbf{Absence}} & \multicolumn{1}{l}{\textbf{Attribute}} & \multicolumn{1}{l}{\textbf{Presence}} & \multicolumn{1}{l}{\textbf{Relation}} & \multicolumn{1}{l}{\textbf{Textual}} & \multicolumn{1}{l}{\textbf{Uniformity}} \\ \midrule
\textit{Open-source Models}\\
Llama3.2 90b  & 92.00 & 83.66 & 87.91 & 88.57 & 89.66 & 86.15 \\
LlavaOV 72b  & 94.34 & 89.57 & 91.49 & 91.16 & 94.51 & 92.75 \\
InternVL2.5 38b  & 94.34 & 90.91 & 90.32 & 87.32 & 93.33 & 91.18 \\
QwenVL2.5 72b  & 94.34 & 90.91 & 91.49 & 90.41 & 90.91 & 87.88 \\
InternVL2.5 78b  & 92.31 & 88.89 & 90.32 & 89.66 & 92.13 & 86.15 \\
\midrule
\textit{closed-source models}\\
GPT-4o  & 98.18 & 94.74 & 94.85 & 91.89 & 92.13 & 94.29 \\
o1 2 & 96.30 & 94.1 & 96.97 & 94.04 & 95.65 & 91.18 \\

Claude  & 94.34 & 87.90 & 92.47 & 88.11 & 90.91 & 76.67 \\
\midrule
Average  & 94.52 &	90.09	& 91.98&	90.15 &	92.40	& 88.28\\
\midrule
Rank  & 1 & 5 & 3 & 4 & 2 & 6 \\ \bottomrule
\end{tabular}%
}
\caption{\textbf{AE performance per category.} AE performance per anomaly category for vanilla inference prompt.}
\label{tab:ae_per_category}
\end{table*}

\subsection{Judge Bias Evaluation}

Since GPT-4o is both one of the evaluated models and the default judge in our LLM-as-a-judge evaluation pipeline for AD, this raises the possibility of bias in its favor. To assess this, we conducted an additional evaluation using Claude-3.5 as an independent judge. Specifically, Claude-3.5 was used to score outputs from both GPT-4o and the top-performing open-source model (InternVL2.5–78B) across several prompting strategies. This cross-model judgment setup allows us to quantify potential self-judging bias and validate the robustness of our conclusions.

\begin{table}[h!]
\centering
\begin{tabular}{@{}lrr@{}}
\toprule
Prompt & \multicolumn{1}{l}{GPT4o} & \multicolumn{1}{l}{InternVL2.5-78B} \\
\midrule
Vanilla & 2.49 & 4.02 \\
CoT & 0.86 & 3.58 \\
CoT + consist. & 1.7 & 4.11 \\
MS CoT & 0 & 0 \\
CoT + SoM & -1.37 & 1.11 \\
\bottomrule
\end{tabular}
\caption{\textbf{Judge bias analysis using Claude-3.5 as an independent judge.} We report average score deltas (Claude minus GPT-4o). Similar positive deltas for GPT-4o and InternVL2.5–78B indicate no self-judging bias. }
\label{tab:claude_judge}
\end{table}

\Cref{tab:claude_judge} reports the delta (Claude judge score minus GPT-4o judge score) for each model and prompt. For most prompts, Claude assigns slightly higher scores than GPT-4o’s self-judgment. Importantly, the deltas are of similar magnitude for both models, and the delta values for GPT-4o are positive, validating the robustness and absence of bias of our evaluation protocol. 

\section{Failure examples}\label{supp:failure} 

\begin{figure*}
    \centering
\begin{promptbox}{Example 1}
\begin{center}
    \includegraphics[width=0.4\linewidth]{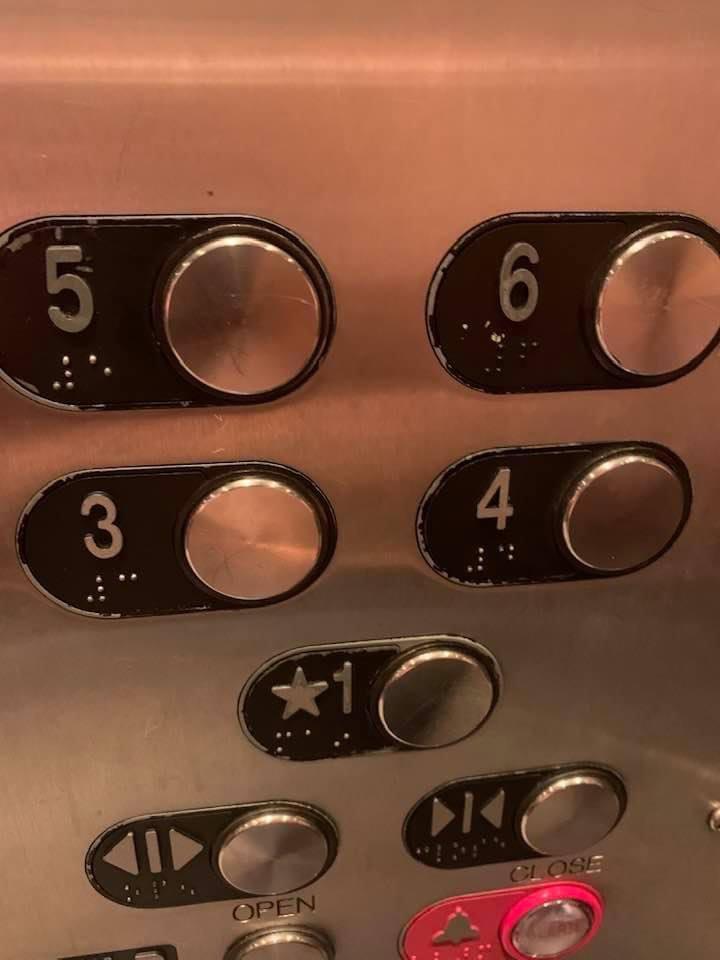}
\end{center}
\textbf{GT AD:} The button for floor number 2 is missing.

\xmark \textbf{o1:} The braille labels for 3 and 5 appear identical., The floor numbers are arranged out of a typical sequence.

\xmark \textbf{GPT-4o:} The button for floor 1 is labeled with a star instead of a number.

\xmark \textbf{LlavaOneVision:} The elevator button for the first floor is marked with a star instead of the number 1., The close button is red, which is unusual as it is typically found in a different color.

\xmark \textbf{InternVL:} The number 1 button is missing its label.

\xmark \textbf{QwenVL:} The button labeled `1' has a star symbol next to it, which is not present on other numbered buttons. 
    \end{promptbox}
\caption{}
\label{fig:ex1}
\end{figure*}

\begin{figure*}
    \centering
\begin{promptbox}{Example 2}
\begin{center}
    \includegraphics[width=0.4\linewidth]{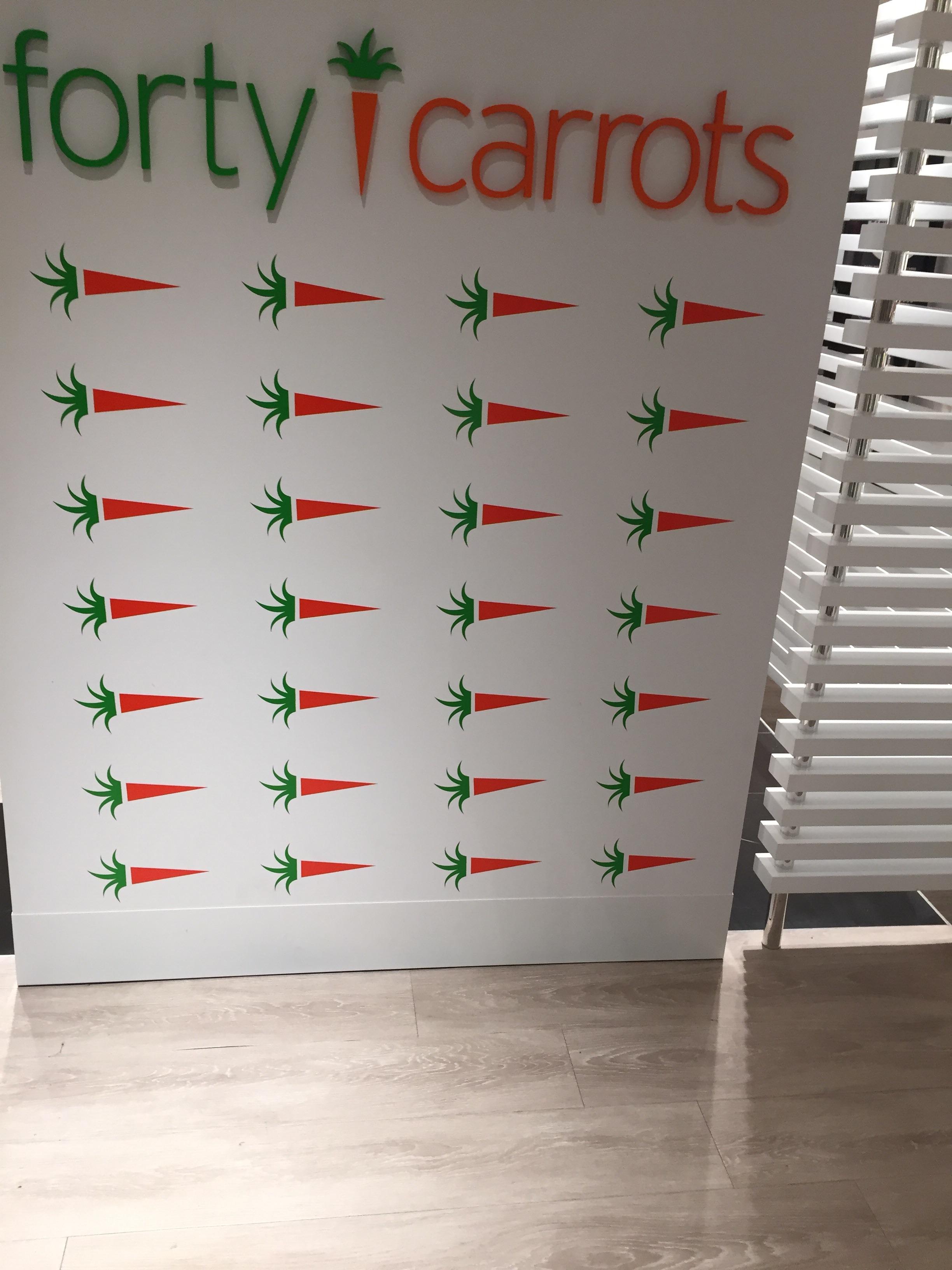}
\end{center}

\textbf{GT AD:} The poster says forty carrots but there are only twenty eight illustrations of carrots.

\xmark \textbf{o1:} No output

\xmark \textbf{GPT-4o:} The word 'forty' is written with a carrot symbol replacing the letter 'i'.

\xmark \textbf{LlavaOneVision:} No output

\xmark \textbf{InternVL:} No output

\xmark \textbf{QwenVL:} No output

\end{promptbox}

\caption{}
\label{fig:ex2}
\end{figure*}

\begin{figure*}
\centering
\begin{promptbox}{Example 3}
\begin{center}
    \includegraphics[width=0.4\linewidth]{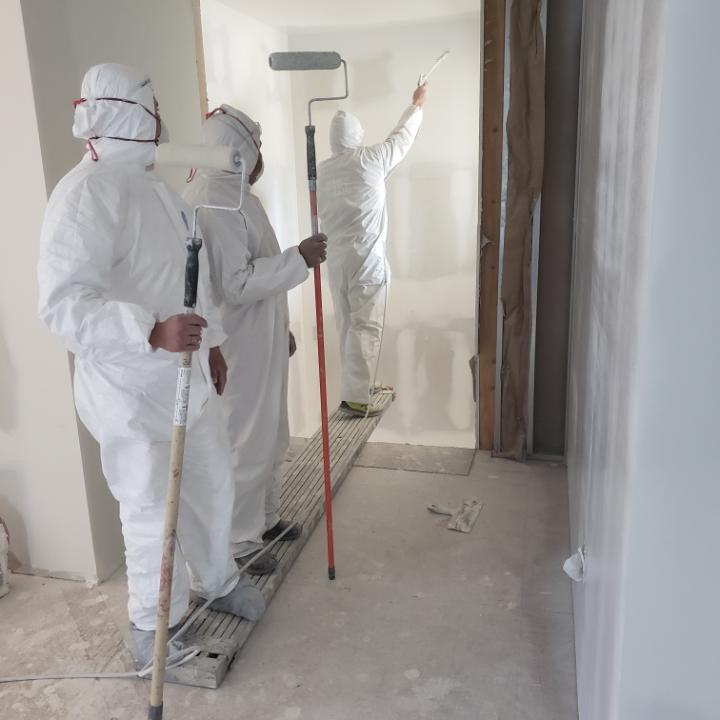}
\end{center}
\textbf{GT AD:} One of the men is standing on the far end of the plank with no other support or safety gear.

\xmark \textbf{o1:} no output

\xmark \textbf{GPT-4o:} no output

\xmark \textbf{LlavaOneVision:} no output

\xmark \textbf{InternVL:} no output

\xmark \textbf{QwenVL:} no output

\end{promptbox}
\caption{}
\label{fig:ex3}
\end{figure*}

\begin{figure*}
\centering
\begin{promptbox}{Example 4}
\begin{center}
    \includegraphics[width=0.4\linewidth]{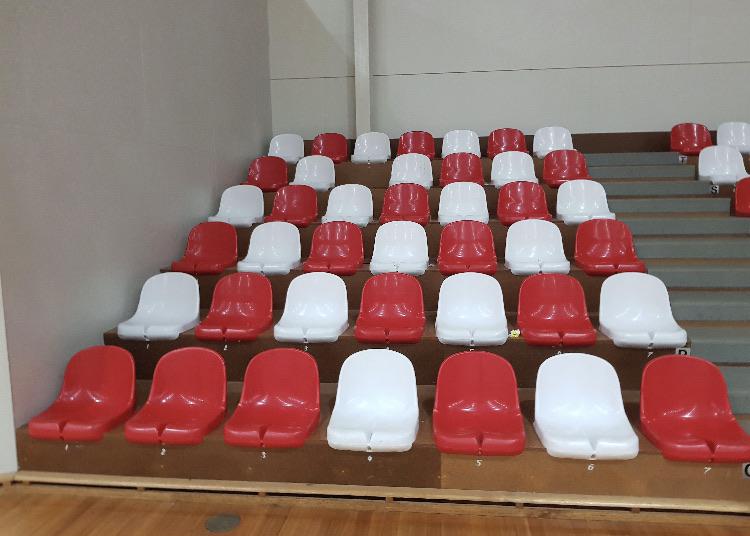}
\end{center}

\textbf{GTAD:} There is a chair missing on the second row from the back.

\xmark \textbf{o1:} No output

\xmark \textbf{GPT-4o:} The seating arrangement is missing a seat in the bottom row, creating an empty space.

\xmark \textbf{LlavaOneVision:} No output

\xmark \textbf{InternVL:} No output

\xmark \textbf{QwenVL:} The seat at the bottom right corner appears to be partially cut off by the edge of the image.

\end{promptbox}
\caption{}
\label{fig:ex4}
\end{figure*}

\begin{figure*}
\centering
\begin{promptbox}{Example 5}
\begin{center}
    \includegraphics[width=0.4\linewidth]{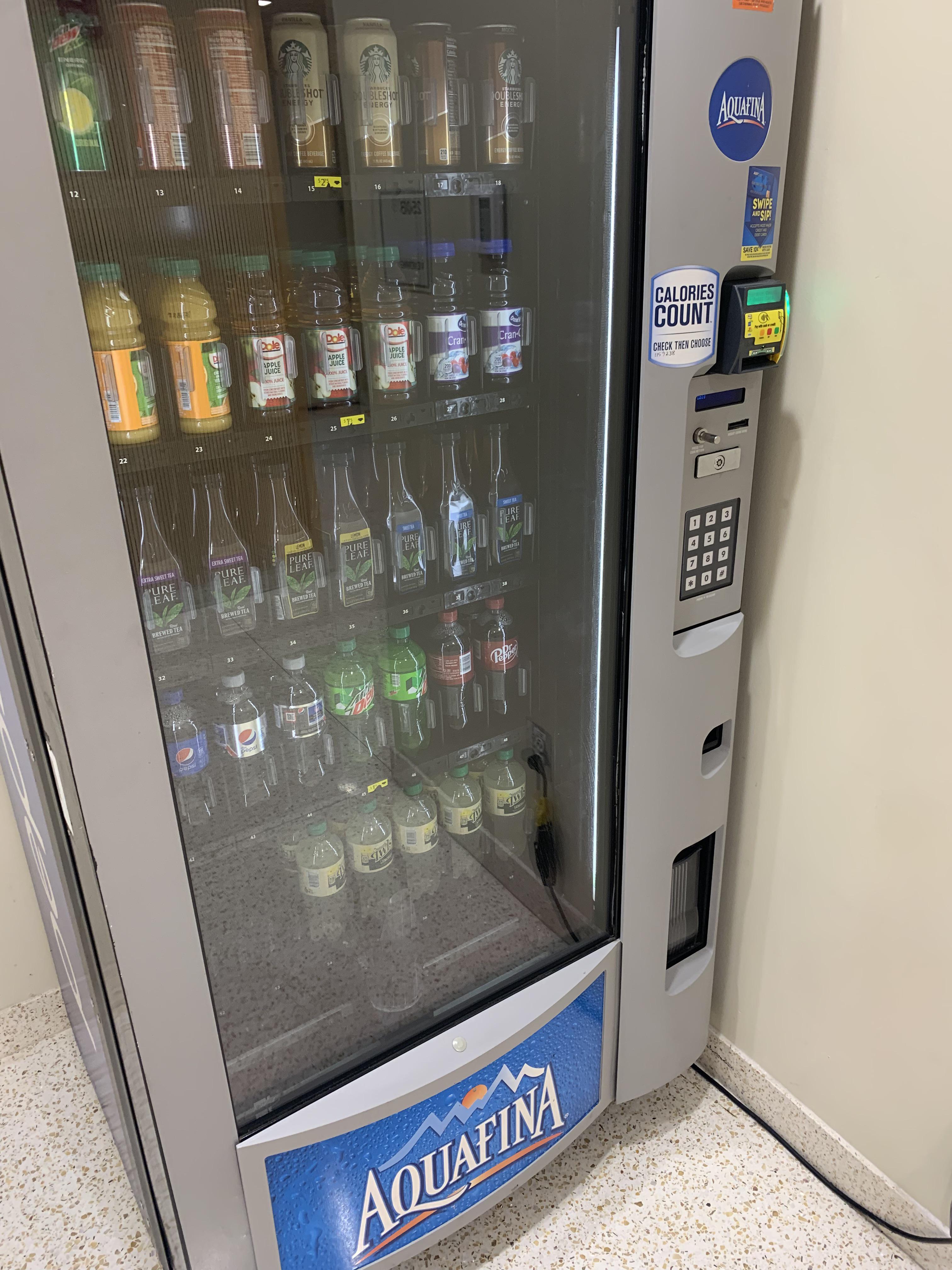}
\end{center}
\textbf{GT AD:} The vending machine is sponsored by Aquafina, a water brand, but there are no Aquafina water bottles being sold by the machine.

\xmark \textbf{o1:} No output

\cmark \textbf{GPT-4o:} The vending machine labeled 'Aquafina' contains a variety of drinks other than Aquafina water.

\xmark \textbf{LlavaOneVision:} No output

\xmark \textbf{InternVL:} A power cord is visible inside the vending machine.

\xmark \textbf{QwenVL:} No output

\end{promptbox}
\caption{}
\label{fig:ex5}
\end{figure*}

\begin{figure*}
\centering
\begin{promptbox}{Example 6}
\begin{center}
    \includegraphics[width=0.4\linewidth]{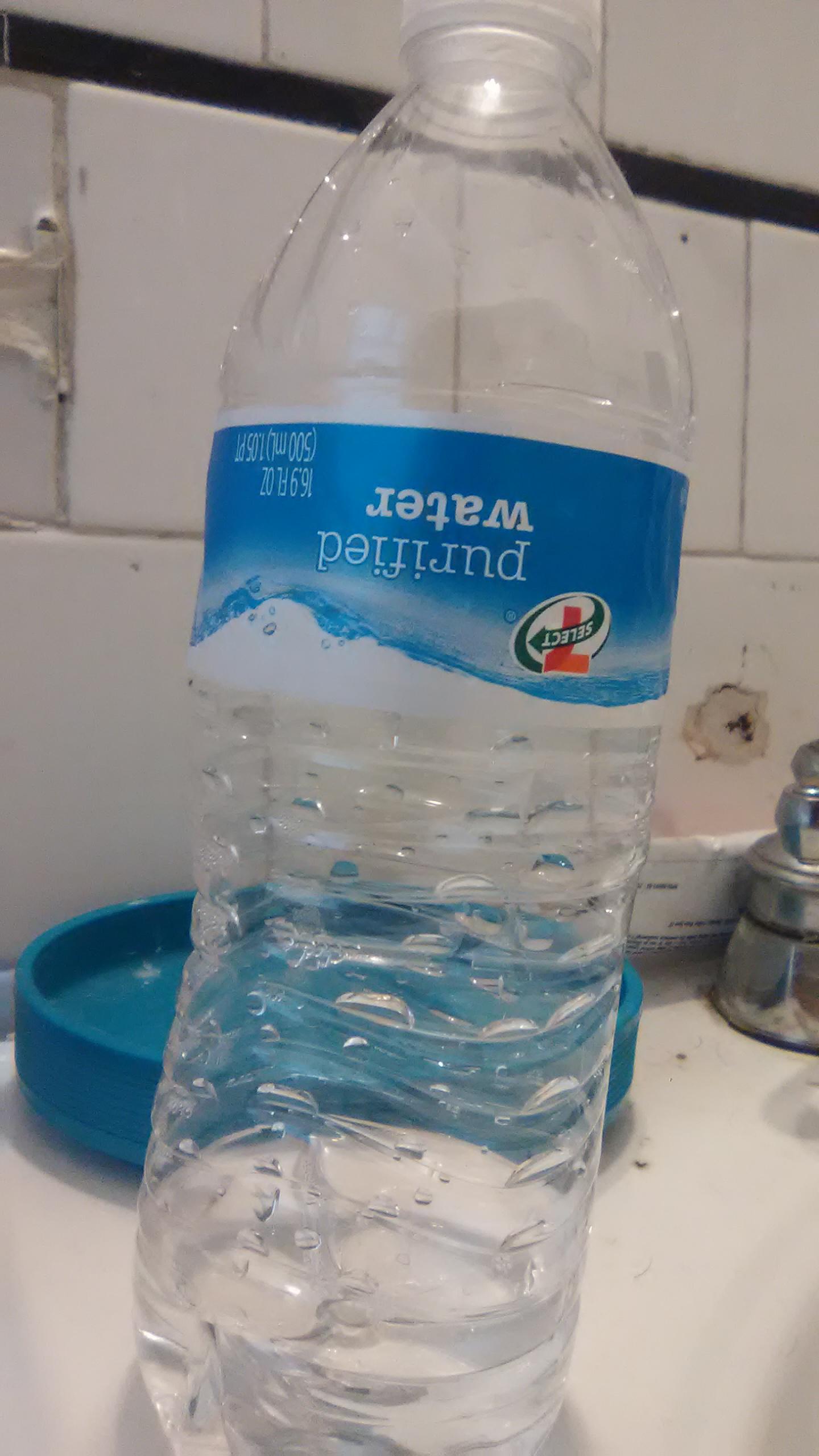}
\end{center}

\textbf{GT AD:} The label on the water bottle is positioned upside down.

\cmark \textbf{GPT-4o:} The label on the water bottle is upside down.

\xmark \textbf{o1:} There is a small hole in the tile with brownish staining surrounding it.

\xmark \textbf{LlavaOneVision:} The water bottle is upside down., The water bottle is not properly sealed.

\cmark \textbf{InternVL:} The label on the water bottle is upside down.

\cmark textbf{QwenVL:} The water bottle label is upside down.

\end{promptbox}
\caption{}
\label{fig:ex6}
\end{figure*}

Examples of model predictions on images of \ourdata{} are shown in \Cref{fig:ex1} to \ref{fig:ex6}.

\begin{figure*}
    \centering
    \includegraphics[width=\linewidth]{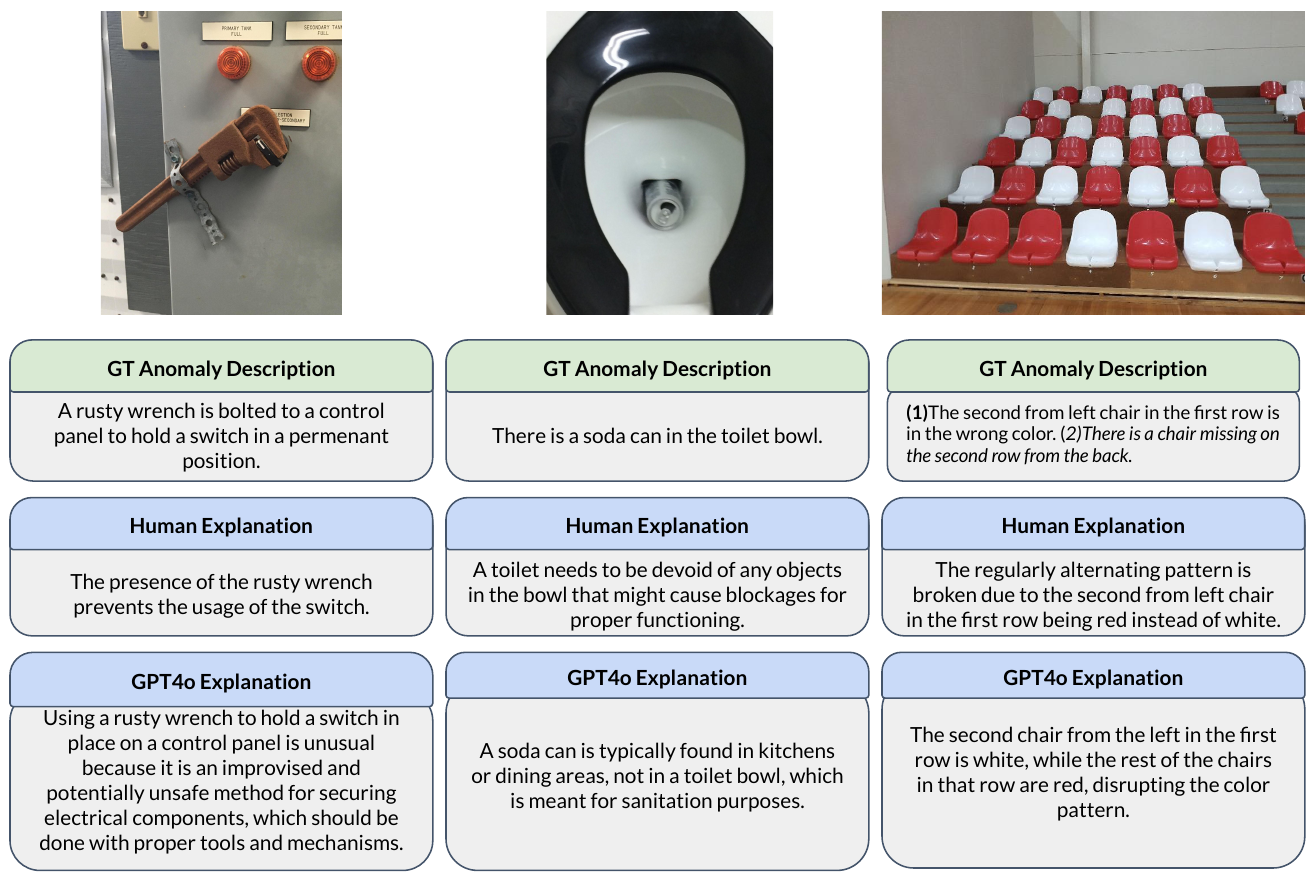}
    \caption{\textbf{GPT-4o Incorrect AE Generations.} Three examples from the AE task where GPT-4o incorrectly explains the anomaly. \textit{Note:} In the rightmost image, two anomalies are present, but only the first one is considered in this example of a failed explanation.}
    \label{fig:ae_error_fig}
\end{figure*}

Examples of incorrect AE generations using GPT-4o are given in \Cref{fig:ae_error_fig}. In the first image, the model generation mentions that the rusty wrench is a functional part of the control panel rather than recognizing it as an improvised obstruction to the switch usage. In the second image, it doesn't give reason for why the toilet bowl should not have objects. In the third image, it misidentifies the anomaly by describing the miscolored chair as white and entirely ignoring the missing chair.

\begin{figure*}
    \centering
    \includegraphics[width=\linewidth]{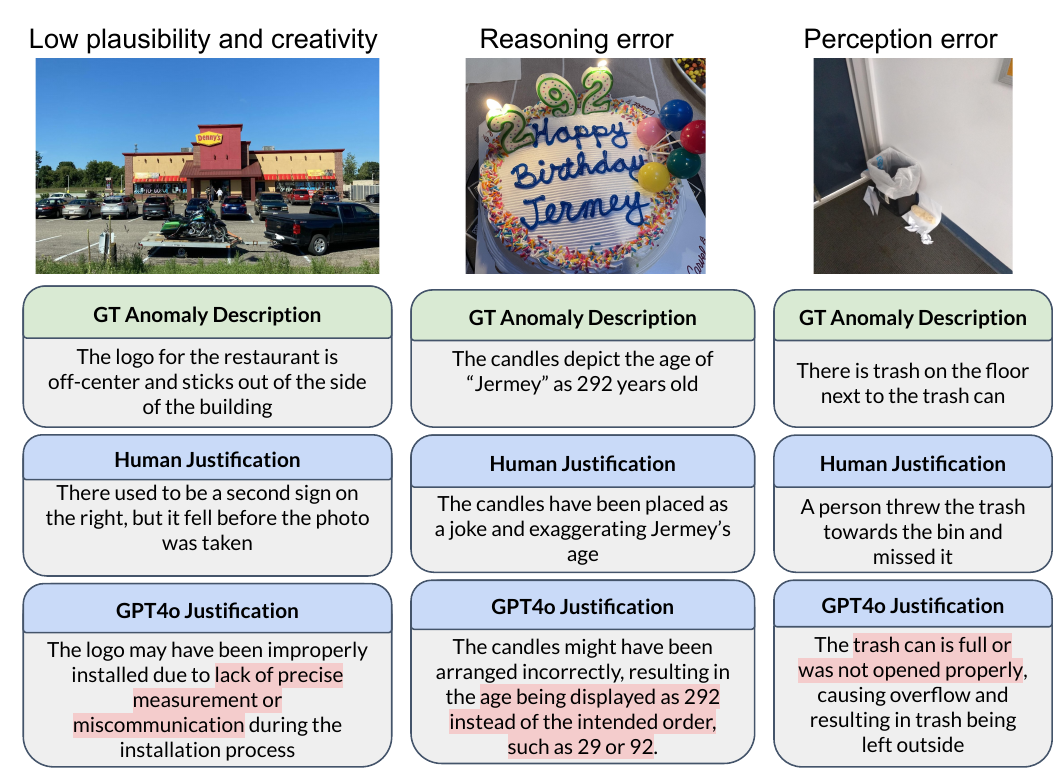}
    \caption{\textbf{GPT-4o Incorrect AJ Generations.} Three examples from the AJ task where GPT-4o provides a poor or incorrect justification.}
    \label{fig:aj_error_fig}
\end{figure*}

Examples of incorrect AJ generations using GPT-4o are given in \Cref{fig:aj_error_fig}. In the first image, the provided justification is highly implausible and lacks creativity. In the second example, the model makes a reasoning mistake when generating the AJ, assuming that 292 is a valid age displayed on a birthday cake. In the third example, the model makes a perception mistake, incorrectly describing the trash can as full or not properly open; the resulting AJ is plausible but incorrect given the image.

\begin{figure*}[h!]
    \centering
\begin{promptbox}{Vanilla Anomaly Description (AD) Prompt}
\textbf{System}: You are an advanced Vision-Language Model designed to extract and describe if any visible anomalies are present in images with precision and brevity.

\textbf{User:} Your task is to accurately describe anomalies observed in images in a concise manner. 

An anomaly is anything that deviates from what most people consider standard, normal, or expected. It can be an unusual element, action, or occurrence in an image that would seem surprising or out of place to most people.

Key principles for anomaly identification:\\
1. Base your descriptions solely on what is clearly visible in the image—no assumptions or external knowledge.\\
2. Descriptions must be objective and straightforward, without explanations, justifications, or subjective language.

Example Format:

If anomalies are present:

\{

    ``anomaly\_description\_1'': [Anomaly description],
    
    ...
    
\}

If no anomalies are visible:

\{\}

Please just provide the description of the anomaly or return `\{\}` if there are no anomalies. **Do not provide any explanation or justification of why it is anomalous.**

Note:
\begin{itemize}[topsep=0pt, itemsep=-0.5em, leftmargin=1em]
\item Limit each description to a maximum of two sentences.
\item Focus on describing the anomaly itself.
\item Do not focus on **why it is unusual or unexpected**.
\end{itemize}
Let's analyze the image and describe the visible anomalies:
\end{promptbox}
\caption{}
\label{fig:AD_prompt}
\end{figure*}

\begin{figure*}[h!]
    \centering
\begin{promptbox}{Chain-of-Thought (CoT) Prompt}
\textbf{System}: You are an advanced Vision-Language Model designed to extract and describe if any visible anomalies are present in images with precision and brevity.

\textbf{User:} 
Your task is to carefully analyze the image, reason step-by-step through what is visible, and then describe any anomalies in a concise and objective manner.

An anomaly is anything that deviates from what most people consider standard, normal, or expected. It can be an unusual element, action, or occurrence in an image that would seem surprising or out of place to most people.

Key principles for anomaly identification:
1. Base your descriptions solely on what is clearly visible in the image—no assumptions or external knowledge.
2. Descriptions must be objective and straightforward, without explanations, justifications, or subjective language.

Example Format:

Let's think step by step: [Thinking steps]

If anomalies are present:

\{

    ``anomaly\_description\_1'': [Anomaly description],
    
    ...
    
\}

If no anomalies are visible:

\{\}

Note:
\begin{itemize}[topsep=0pt, itemsep=-0.5em, leftmargin=1em]
\item Limit each description to a maximum of two sentences.
\item Focus on describing the anomaly itself.
\item Do not focus on **why it is unusual or unexpected**.
\end{itemize}

Let's analyze the image, think step by step and then describe the visible anomalies:

\end{promptbox}
\caption{}
\label{fig:AD_cot_prompt}
\end{figure*}

\begin{figure*}[h!]
    \centering
\begin{promptbox}{Multi-step reasoning (MS CoT) Prompt}
\textbf{System}: You are an advanced Vision-Language Model designed to extract and describe if any visible anomalies are present in images with precision and brevity.

\textbf{User:} Your task is to accurately describe anomalies observed in images in a concise manner. 

An anomaly is anything that deviates from what most people consider standard, normal, or expected. It can be an unusual element, action, or occurrence in an image that would seem surprising or out of place to most people.

Your goal is to carefully analyze the image using simple, structured reasoning, and describe any visible anomalies. Do not use external knowledge or assumptions — only what can be clearly seen in the image.

Use the following structure in your response:

1. **Planning**: Briefly explain the steps you will take to perform the task.

2. **Image Contents**: List the main elements visible in the image (e.g. objects, people, actions, text).

3. **Step-by-step reasoning**: Think through the image in a logical sequence to identify if anything looks unusual or out of place.

4. **Final Answer**:
If anomalies are present:

\{

    ``anomaly\_description\_1'': [Anomaly description],
    
    ...
    
\}

If no anomalies are visible:

\{\}

Note:
\begin{itemize}[topsep=0pt, itemsep=-0.5em, leftmargin=1em]
\item Limit each description to a maximum of two sentences.
\item Focus on describing the anomaly itself.
\item Do not focus on **why it is unusual or unexpected**.
\end{itemize}

Let's begin by planning, then analyzing the image step by step, and finally reporting any anomalies found:

\end{promptbox}
\caption{}
\label{fig:AD_msr_prompt}
\end{figure*}

\begin{figure*}[h!]
    \centering
\begin{promptbox}{Self-consistency ensembler Prompt}
\textbf{System}: You are an advanced Vision-Language Model designed to extract and describe if any visible anomalies are present in images with precision and brevity.

\textbf{User:} 

You are given three sets of anomaly descriptions for the same image:
\\
1. [\textit{Anomaly Descriptions from 1st inference}]

2. [\textit{Anomaly Descriptions from 2nd inference}]

3. [\textit{Anomaly Descriptions from 3rd inference}]
\\
Your job is to identify the anomaly descriptions that are repeated — that is, descriptions that appear in at least twice. These may be worded slightly differently but must describe the same anomaly.

Do not make up any new descriptions. Ignore differences in phrasing if the meaning is clearly the same.

Return only the repeated anomaly descriptions in the given json format. 

**Only include those that appear atleast twice.**

Example Output Format:

\{

    ``anomaly\_description\_1'': [Anomaly description],
    
    ...
    
\}

Do not include any extra explanation.

\end{promptbox}
\caption{}
\label{fig:AD_self_consistency_prompt}
\end{figure*}

\begin{figure*}[h!]
\centering
\begin{promptbox}{Anomaly Localization Prompt}
\textbf{System:}
You are an advanced Vision-Language Model designed to locate the given anomaly description.\\
\textbf{User:} 
Your task is to find and localize the single anomaly described below in the given image.\\

Anomaly Description:  \{anomaly\_description\}\\

Image resolution: \{width\} x \{height\} pixels.\\

Please provide your output in JSON format exactly as follows:

\{\{\\
"box":
["x1": <top-left x>,  "y1": <top-left y>,\\  "x2": <bottom-right x>,  "y2": <bottom-right y>
]
\\\}\}
    
\end{promptbox}
 \caption{}
 \label{fig:AL_prompt}
\end{figure*}

\begin{figure*}
\begin{promptbox}{Anomaly Explanation (\texttt{AE}) Prompt}
\textbf{System}: You are an advanced Vision Language Model designed to analyze anomalies seen in images.

\textbf{User:} You will be provided with an image along with a textual description of an anomaly present in the image. 
Explain why the anomaly is anomalous.

Anomaly: [Anomaly description and explanation]

Let’s analyze the image and the anomaly first and then provide a concise response (max 2 sentences) in a json format:

\{

    ``explanation'': [Explain why it is unusual or unexpected]
    
\}

\end{promptbox}
\caption{}
\label{fig:AE_prompt}
\end{figure*}

\begin{figure*}
\begin{promptbox}{Anomaly Justification (\texttt{AJ}) Prompt}
\textbf{System}: You are an advanced Vision Language Model designed to analyze anomalies seen in images.

\textbf{User:} You will be provided with an image and the description of an anomaly seen in the image. Provide a single, most realistic and plausible reason for how the anomaly happened.

Anomaly: [Anomaly description and explanation]

Let's analyse the image and the anomaly first and then provide a concise response (max 2 sentences) in a json format:

\{

    ``justification'': [Describe how the anomaly happened]
    
\}
\end{promptbox}
\caption{}
\label{fig:AJ_prompt}
\end{figure*}
\begin{figure*}
\begin{promptbox}{Anomaly Severity Prompt}
\textbf{System}: You are an advanced Vision-Language Model designed to analyze images and detect anomalies.

\textbf{User:} Assess the severity of the consequence of the anomaly on a scale from 1 to 5:
\begin{enumerate}[topsep=0pt, itemsep=-0.5em, leftmargin=1em]
    \item (No consequence): The anomaly is purely aesthetic or has no functional/safety impact. Example: A tile of a different color on the pavement.
    \item (Low Concern)
    \item (Moderate Concern): The anomaly may cause inconvenience or inefficiency but does not pose immediate risks. Example: A misaligned sign that is still readable.
    \item (High Concern)
    \item (Requires Immediate Action): The anomaly presents a safety hazard, major malfunction, or significant risk. Example: A worker using a circular saw without protection gear.
\end{enumerate}

\textbf{Inputs:}

- Image: (Attached image)

- Anomaly Description: \{\}\newline

Provide a severity rating in this format:

\{

    ``severity'': [Score between 1 and 5]
    
\}
\end{promptbox}
\caption{}
\label{fig:severity_prompt}
\end{figure*}

\begin{figure*}
\begin{promptbox}{Anomaly Surprisal Prompt}
\textbf{System}: You are an advanced Vision-Language Model designed to analyze images and detect anomalies.

\textbf{User:} Assess how surprising or uncommon the anomaly is on a scale from 1 to 5:
\begin{enumerate}[topsep=0pt, itemsep=-0.5em, leftmargin=1em]
    \item (Common): Frequently observed in similar contexts; most people would not be surprised. Example: A car parked in an inconvenient way.
    \item (Relatively Common)
    \item (Average): Might raise curiosity but not shock. Example: A person eating spaghetti with chopsticks.
    \item (Uncommon)
    \item (Extremely Rare): Highly uncommon and surprising; most people have never seen it before. Example: A tree growing upside down from a roof.
\end{enumerate}

\textbf{Inputs:}

- Image: (Attached image)

- Anomaly Description: \{\}\newline

Provide a surprisal rating in this format:

\{

    ``surprisal'': [Score between 1 and 5]
    
\}
\end{promptbox}
\caption{}
\label{fig:sup_prompt}
\end{figure*}

\begin{figure*}
\begin{promptbox}{Anomaly Complexity Prompt}
\textbf{System}: You are an advanced Vision-Language Model designed to analyze images and detect anomalies.

\textbf{User:} Assess how difficult it would be for a person to detect the anomaly on a scale from 1 to 5:
\begin{enumerate}[topsep=0pt, itemsep=-0.5em, leftmargin=1em]
    \item (Easy): Most people would notice the anomaly immediately without effort. Example: A red apple among green apples.
    \item (Mild)
    \item (Moderate): Requires some focus to identify but becomes clear after a few seconds. Example: A misspelled word on a sign.
    \item (Difficult)
    \item (Very difficult): Blends into the surroundings or demands specific knowledge to identify. Example: A contradiction in the screenshot of an email.
\end{enumerate}

\textbf{Inputs:}

- Image: (Attached image)

- Anomaly Description: \{\}\newline

Provide a complexity rating in this format:

\{

    ``complexity'': [Score between 1 and 5]
    
\}
\end{promptbox}
\caption{}
\label{fig:comp_prompt}
\end{figure*}


\begin{figure*}
\begin{promptbox}{Anomaly Description Evaluation Prompt}
\textbf{System}: You are an advanced AI assistant designed to compare two descriptions of an anomaly in the image attached.

\textbf{User:} Compare the following two descriptions of an anomaly in an image. Judge whether they describe the same anomaly. If they match, respond with 'Yes' and briefly explain why. If they differ, respond with 'No' and provide a reason for the difference.

REFERENCE: [\textit{Ground truth anomaly description}]

RESPONSE: [\textit{Model-generated anomaly description}]
\end{promptbox}
\caption{}
\label{fig:ad_judge_prompt}

\end{figure*}

\begin{figure*}
\begin{promptbox}{Anomaly Explanation Evaluation Prompt}
\textbf{System:} You are an advanced AI assistant designed to compare two explanations for a visual anomaly.

\textbf{User:} Determine whether the model explanation accurately reflects the core reasoning in the human annotation for why the given anomaly is considered unusual in the image. 

The explanation does not need to match the human annotation word-for-word, but it should be logically aligned and refer to the same underlying cause.

Minor differences in wording are acceptable, but explanations that are unrelated or based on a different logic should be marked as incorrect.

Anomaly Description: [\textit{Ground truth anomaly description}]

Human explanation: [\textit{Human annotation}]

Model explanation: [\textit{Model-generated anomaly explanation}]

If the explanations are unrelated or based on a different logic, answer 'False'.
\end{promptbox}
\caption{}
\label{fig:ae_judge_prompt}
\end{figure*}

\begin{figure*}
\begin{promptbox}{Anomaly Category Classification Prompt}
\textbf{System:} You are an expert in classifying visual anomalies based on descriptions.

\textbf{User:} You are given a taxonomy of anomaly types:

1. Entity Presence –An object is present when it shouldn't be.

2.Entity Absence – An expected object is missing.

3. Entity Attribute – An object has an unusual attribute (color, shape, label, orientation, usage).

4. Spatial Relation – Objects are positioned or oriented incorrectly relative to one another.

5. Uniformity Breach – A disruption in an expected pattern or symmetry.

6. Textual Anomaly – The image contains text that is contradictory, unexpected, or illogical.
\\
Given the following anomaly description, classify it into one of the five categories. Only respond with the category name.\\

Anomaly description: [\textit{Model generated anomaly description}]

\end{promptbox}
\caption{}
\label{fig:fp_classifier_prompt}

\end{figure*}

\begin{figure*}
\begin{promptbox}{Anomaly Cultural Analysis Prompt}

\textbf{User:} You are a culturally-aware AI with expertise in global customs, social norms, and visual analysis. Based on the image, description, and noted anomaly:

Analyze the anomaly within its cultural context.

Determine if it aligns with any specific cultural, religious, regional, or historical norms.

If yes, identify the culture/region and explain why this is considered normal there.

If no, clearly state that and briefly explain why it does not align culturally.

Be objective, respectful, and avoid stereotypes. Consider that some anomalies may have universal meaning without cultural bias.

Respond as a dictionary with keys:

- cultural alignment: ``yes'' or ``no''

- context: the relevant cultural norm that explains the anomaly (or null if none)

- justification: explanation why the anomaly is normal or not culturally aligned

\textbf{Inputs:}

- Image: (Attached image)

- Anomaly Description: [\textit{Ground truth anomaly description}]

\end{promptbox}
\caption{}
\label{fig:cultural_bias_prompt}
\end{figure*}
\end{document}